%% file: main.tex
\pdfoutput=1 

\documentclass[11pt]{article}

\usepackage[pdftex]{color,graphicx}
\usepackage{amssymb}
\usepackage{subcaption}
\usepackage{soul}
\usepackage[normalem]{ulem}
\usepackage[ruled, vlined, linesnumbered]{algorithm2e}

\newcommand{\supe}{\textcolor[rgb]{0.0,0.7,0.0}{$\blacktriangle$}}
\newcommand{\infe}{\textcolor[rgb]{1.0,0,0.0}{$\blacktriangledown$}}
\newcommand{\emp}{\textcolor[rgb]{1.0, 0.49, 0.0}{$\bullet$}}

\usepackage[final]{acl}

\usepackage{times}
\usepackage{latexsym}

\usepackage[T1]{fontenc}

\usepackage[utf8]{inputenc}

\usepackage{microtype}

\usepackage{inconsolata}

\title{A Strategy to Combine 1stGen Transformers and  Open LLMs for Automatic Text Classification}

\author{Claudio M. V. de Andrade$^1$, Washington Cunha$^1$, 
Davi Reis$^2$, Adriana Silvina Pagano$^1$, \\ {\bf Leonardo Rocha}$^2$, {\bf Marcos André Gonçalves}$^1$ \\ (claudio.valiense, washingtoncunha, mgoncalv)@dcc.ufmg.br, \\ davireisjesus@aluno.ufsj.edu.br, apagano@ufmg.br, lcrocha@ufsj.edu.br \\
$^1$ Federal University of Minas Gerais, Brazil\\ $^2$ Federal University of São João Del-Rei, Brazil
 }

\begin{document}
\maketitle

\input{sections/0_abstract}
\input{sections/1_introduction}
\input{sections/2_related_work}

\input{sections/3_cmbb}
\input{sections/4_methodology}
\input{sections/5_results}
\input{sections/6_conclusion}
\newpage
\input{sections/7_limitation}


\bibliography{main}
\input{sections/8_appendix}

\end{document}

%% file: sections/0_abstract.tex
\begin{abstract}
Transformer models have achieved state-of-the-art results, with Large Language Models (LLMs), an evolution of first-generation transformers (1stTR), being considered the cutting edge in several NLP tasks. However, the literature has yet to conclusively demonstrate that LLMs consistently outperform 1stTRs across all NLP tasks. This study compares three 1stTRs (BERT, RoBERTa, and BART) with two open LLMs (Llama 2 and Bloom) across 11 sentiment analysis datasets. The results indicate that open LLMs may moderately outperform or match 1stTRs in 8 out of 11 datasets but only when fine-tuned. Given this substantial cost for only moderate gains, the practical applicability of these models in cost-sensitive scenarios is questionable. In this context, a confidence-based strategy that seamlessly integrates 1stTRs with open LLMs based on prediction certainty is proposed. High-confidence documents are classified by the more cost-effective 1stTRs, while uncertain cases are handled by LLMs in zero-shot or few-shot modes, at a much lower cost than fine-tuned versions. Experiments in sentiment analysis demonstrate that our solution not only outperforms 1stTRs, zero-shot, and few-shot LLMs but also competes closely with fine-tuned LLMs at a fraction of the cost.
\looseness=-1
\end{abstract}

%% file: sections/1_introduction.tex
\section{Introduction}

Automatic text classification (ATC) is essential in multiple contexts to enhance efficiency and effectiveness in tasks ranging from organizing large volumes of data to personalizing user experiences. ATC has experienced a huge revolution recently with the emergence of semantically enriched Transformer models ~\cite{Devlin_2018} which achieve state-of-the-art performance in most ATC scenarios~\cite{artigo_representacao,cunha2023effective,zanotto2021stroke, pasin2024quantum}.\looseness=-1   

In recent years, Large Language Models (LLMs) have arisen, being built on top of first-generation transformers (1stTRs for short). Works have pointed to these new models as the current SOTA for some NLP tasks~\cite{holistic}. Although the literature reports LLMs' superiority for specific tasks, such as summarization and translation, for others, such as sentiment analysis (our focus in this paper), it is still not clear if LLMs' complexity and size (e.g., in terms of parameters) translate into statistical and mainly, \textit{practical} gains. For instance, several studies point RoBERTa, a 1stTR representative,  as a very strong model for sentiment analysis~\cite{cunha23csur}, ranking prominently on leaderboards such as the GLUE benchmark\footnote{\url{https://gluebenchmark.com/leaderboard/}}.

In light of the above discussion, the first research question we aim to answer in this paper is {\bf RQ1:} ``Are (open) LLMs more effective in overcoming the limits of first-generation transformers (in sentiment analysis)?''. \citet{survey} indicate there is no consensus that LLMs always perform better in classification. Thus, to address this question, we ran a comprehensive set of experiments comparing these two generations, using three popular 1stTR representatives  (BERT, RoBERTa, and BART) and two open LLMs (Llama 2 and Bloom) using a benchmark composed of 11 sentiment analysis datasets with different characteristics. In particular, two of these datasets have been collected after the release of the LLMs (IMDB2024, RottenA2023) to minimize potential data contamination issues~\cite{holistic}. In this comparison, we focus on open-source LLMs, as closed-source and proprietary LLMs, such as ChatGPT, are black boxes that prevent us from understanding how they were trained or their internal structure\footnote{Closed LLMs are irreproducible~\cite{gao2024llm}.}. Our results indicate that, indeed, in most cases (but not all), open LLMs can (moderately)  overcome 1stTRs.\looseness=-1 

Depending on the training (or its absence),  LLM approaches can be divided into three groups: zero-shot, few-shot, and fine-tuning (aka as full-shot\footnote{We use the terms full-shot and fine-tuning 
as synonyms.\looseness=-1
}). In a zero-shot approach, the model is expected to perform tasks without any specific training on those tasks. In a few-shot approach, the model is given a small number of examples to learn from before performing the task. Finally, fine-tuning is supervised and applied to a domain-specific labeled dataset, allowing for further model optimization. The observed gains of open LLMs over 1stTRs, reaching up to 7.5\%, are obtained mostly in the \textbf{full-shot} scenario at the \textbf{very high cost} of fine-tuning the LLMs. Indeed, as we shall discuss, zero-shot or even few-shot versions of LLMs are not able to outperform fine-tuned 1stTRs in several cases.\looseness=-1

Given the (much) higher computational costs associated with running open  LLMs in their most effective form (full-shot strategy) to obtain small to moderate gains over 1stTRs, a natural (research) question that arises is {\bf RQ2:} ``What is the (computational) cost of using open LLMs for ATC in comparison to 1stTR cost?''. To answer this question, we conducted a thorough analysis of our experimental results, considering zero-shot, few-shot, and fine-tuning strategies for the LLMs to assess the tradeoffs between effectiveness and costs in terms of computational time to train the models and their impact on carbon emission. We found that LLMs are orders of magnitude more costly to fine-tune when compared to 1stTRs --  full-shot and few-shot LLMs are up to 13x and 1.7x slower, respectively, than the 1stTRs.\looseness=-1

As current open LLMs can produce moderate gains over 1stTRs only through
highly costly fine-tuning processes, depending on the application scenario, the benefits may not be worth the costs. For instance, in scenarios where computational costs must be considered, such as a virtual assistant for medical assistance, the assistant cannot afford to be slow in responding or ineffective~\cite{biswas2023can}. Other examples include 
streaming platforms, where data are updated constantly, making retraining the model multiple times impractical. These are situations where the computational cost needs to be carefully weighed~\cite{10.1145/3404835.3462919}.\looseness=-1

All this leads to our final research question {\bf RQ3:} ``Is it possible to perform a combination of 1stTRs and (open) LLMs in order to achieve a better effectiveness/cost tradeoff when compared to these options separately?'' To answer this question, we propose a novel confidence-based strategy called ``Call My Big Sibling'' (\textbf{CMBS}) that smoothly combines 1stTRs and (open) LLMs \textbf{based on uncertainty}. In CMBS, we rely first on \textit{fine-tuned 1stTR models}, which are already very effective and efficient, to generate contextual representations for each document and use them along with \textbf{calibrated} classifiers\footnote{In a calibrated classifier, the probability of the output directly correlates with the accuracy of the classifier, allowing to interpret the probability as confidence on the prediction.} to obtain a \textit{reliable} confidence score on the prediction. We chose to use calibrated classifiers instead of the original softmax-based classification layer of 1stTRs\footnote{Recent work has shown that modifying the classification layer does not affect the effectiveness of the Transformer~\cite{Andrade21}.} so that we can trust their confidence (i.e., the certainty of the classifier on the prediction)  -- softmax-based Transformers are known to be uncalibrated~\cite{WenlongICLR2024}.\looseness=-1

In sum, high-certainty documents (i.e., with high certainty scores) are classified by fine-tuned 1stTR models, while low-certainty documents 
are sent to the \textit{zero-shot or few-shot versions of the LLMs} for classification. By applying zero-shot, we incur no further training (or tuning) costs for our combined solution. The combination with few-shot incurs just a little extra time to tune the LLMs with a few examples, but with potentially additional gains in effectiveness, which is an attractive, cost-effective option in some cases, as we shall discuss. 

Our experiments demonstrate the practical effectiveness of our CMBS solution. It outperforms the best 1stTRs in 7 out of 11 datasets (tying in the other 4), with a marginal increase in computational cost over 1stTRs of approximately 14\% for zero-shot and 23\% for few-shot. Moreover, in 9 of the 11 datasets, CMBS is better or ties with the best fine-tuned LLM at a fraction (1/13 on average) of the cost of fine-tuning the LLM (full-shot).\looseness=-1

In sum, the main contributions of this paper are:

\begin{itemize}
\vspace{-0.3cm}
    \item We perform a comprehensive comparative evaluation of the cost/effectiveness  tradeoffs between 1stTRs and (open) LLMs.\looseness=-1 

\vspace{-0.3cm}
    \item We propose ``Call My Big Sibling'' (\textbf{CMBS}), a confidence-based strategy to combine 1stTR and LLMs that achieve the best cost/effectiveness tradeoff considering three 1stTRs (BERT, RoBERTa, and BART) and the zero-shot, few-shot and full-shot versions of two open LLMs - Llama2 and Bloom.\looseness=-1
\end{itemize}

%% file: sections/2_related_work.tex
\section{Related Work}
\label{trabalhos_relacionados}

LLMs' computational costs have led to numerous studies highlighting their financial and environmental impacts. For instance, \citealp{strubell2019energy} illustrates the substantial financial costs propelled by the continuous need for investment in specialized hardware to manage progressively larger language models. This trend not only limits access to these models but also escalates energy consumption, affecting the environment by increasing carbon dioxide (CO$_2$) emissions.\looseness=-1

Among LLMs, there are proprietary and closed-source ones, such as \textit{chatGPT}, which operate as black boxes. This opacity poses challenges in comprehending their training methodologies or internal structures, thereby obstructing reproducibility in research reliant on these models. Moreover, utilizing such LLMs often entails transmitting data through web platforms or APIs, a delicate issue when data is sensitive and cannot be shared. As a result, numerous studies advocate for restricting scientific evaluations to run locally, open-source LLMs such as Bloom and Llama 2~\cite{spirling2023open}, \looseness=-1

\citealp{holistic} investigate various LLMs across multiple tasks, prompts, metrics, and datasets. Like Liang et al., we include LLM evaluation and the tradeoff between efficiency and effectiveness. Unlike their study, which focuses on the breadth of evaluation with several domains (including only one sentiment dataset), our work is depth-oriented into the specific task of \textbf{sentiment classification}, covering multiple datasets with diverse characteristics and domains. Moreover, although Lian et al. evaluate several models, they do not compare them with 1stTRs such as RoBERTa, considered SOTA in sentiment classification~\cite{roberta_forte, cunha2021cost,cunha2020extended,francca2024representation,belem2024novel}. As we shall see, some 1stTRs do outperform LLMs in sentiment datasets. Finally,  they do not provide any solution for the effectiveness-cost tradeoff problem. \textbf{\textit{We do!}}\looseness=-1

%% file: sections/3_cmbb.tex
\section{CMBS Proposal}
\label{CMBS}

One of the main contributions of our work is the proposal of a novel strategy to combine simpler, more efficient, but perhaps less effective 1stTRs with potentially more effective but costly LLMs, aiming to promote effectiveness while minimizing computational cost. Our solution ``Call-My-Big-Sibling'' (CMBS), metaphorically conjures up the image of a small (but smart) child who, in a challenging situation, seeks help from a bigger sibling,  pursues the best tradeoff between effectiveness and cost with a confidence-based pipeline of Transformers (1stTRs -> LLMs).\looseness=-1

CMBS seamlessly integrates 1stTRs and (open) LLMs by leveraging uncertainty. In this framework, we first employ \textit{fine-tuned 1stTR models}\footnote{Tuned with the full training data.}, which are already highly effective in sentiment classification (and faster to tune compared to LLMs), to create contextual representations for each document. These representations are then utilized to train \textbf{calibrated} classifiers capable of deriving a confidence score for the prediction of each document (at test time). Rather than utilizing the original softmax-based classification layer of 1stTRs, we opt for using more calibrated classifiers\footnote{\cite{artigo_representacao} has shown that substituting classifiers has no effect at all in effectiveness.}. This choice enables us to place greater trust in the confidence levels of the classifiers. It is worth noting that while Transformers are recognized for their effectiveness, they are also acknowledged for being uncalibrated \cite{WenlongICLR2024}. In our solution, (test) documents classified below a certain confidence threshold (a method parameter) by the 1stTRs are sent to an open LLM to be classified.\looseness=-1

\vspace{-0.3cm}
\begin{figure}[!h]
  \centering
  \includegraphics[width=0.45\textwidth]{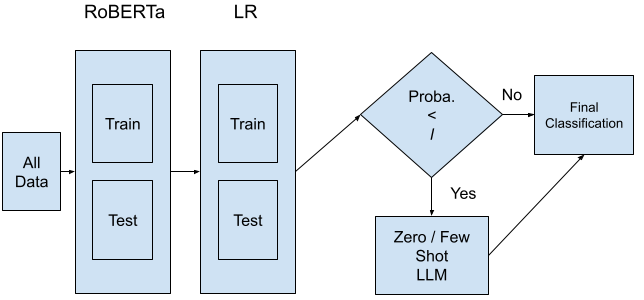}
  \vspace{-0.3cm}
  \caption{\footnotesize{Flowchart of the Evaluation Methodology.}}
  \label{fig:fluxograma}
\end{figure}
\vspace{-0.3cm}

The procedure is illustrated in Figure~\ref{fig:fluxograma}\footnote{The CMBS pseudocode can be seen in Appendix~\ref{appendix:pseudocode}.}. First, we obtain (fine-tunned) RoBERTa representations for all documents. In this representation, the ``CLS'' token is utilized in the model to represent the entire document, corresponding to a pooling calculated using self-attention rather than average or max-pooling, with the vectors of the document tokens.\looseness=-1

Second, we use these representations to train a Logistic Regression model that predicts calibrated probabilities for each class. Calibrated models offer a powerful approach to text classification, providing more reliable predictions and deeper insights into the uncertainty of predictions. At prediction (test) time, based on the LR output probabilities, we select a set for which the classifier is most uncertain about the classification to send to a large model (LLM) for final prediction. Due to the computational cost associated with full-shot second-generation transformers, we employ either the zero-shot or few-shot strategies for these LLMs. Finally, our final prediction set is built using the following procedure: 1) we evaluate the probability provided by the calibrated LR model and compare it with the threshold parameter; 2) we then decide whether the prediction will be made using a 1stTR or an LLM (zero or few-shot). The pseudo-code for the whole procedure can be found in Appendix A.\looseness=-1

%% file: sections/4_methodology.tex
\section{Experimental Methodology and Setup}
\label{metodologia}

\subsection{Datasets}

Our study draws on \textbf{eleven} datasets developed for binary sentiment classification. 
The datasets include \textbf{Finance}, \textbf{IMDB}, \textbf{PangMovie}, \textbf{SemEval2017}, \textbf{SiliconeMeldS}, \textbf{SiliconeSem}, \textbf{SST}, \textbf{SST2} and \textbf{Yelp Review}. Moreover, with the significant amount of data used in building LLMs, several authors express concerns about contamination in evaluation data. Intending to minimize this issue, we collected and curated two datasets with data post-LLMs release (\textbf{RottenA2023} from August 2023 to February 2024 and \textbf{IMDB2024} from January 2024 to May 2024), ensuring no contamination in the training of these LLMs. Web scraping was performed using Beautiful Soup~\cite{hajba2018using}, a popular Python library. Several works in the text classification field have used most of these datasets as benchmarks. See Appendix~\ref{appendix:datasets} for further information about the datasets, including aspects such as domain, number of documents, density and skewness (class imbalance). As we can see, our benchmark covers a high variety of heterogeneous aspects.  \looseness=-1

\subsection{Data Representation}
\label{data_representation}

We fine-tuned BERT, RoBERTa, and BART, adapting these 1stTRs to the specific domain of sentiment classification using the texts and labels in our datasets. The aim was to improve the representation and enhance the model's effectiveness for sentiment classification. The model's fine-tuning produces an encoder, which generates CLS-based 768-dimensional embedding vectors to represent the documents. As discussed in~\cite{artigo_representacao,cunha2023tpdr}, this fine-tuning process is fundamental to ensure the quality of the representation and the separability (into semantic classes) of the generated embedding space.\looseness=-1 

\subsection{Prompt Template}

We investigated the performance of two open LLMs (Llama2 and Bloom) optimizing each LLM through prompt-learning  ~\citep{luo2022prcbert,info14050262}. This method entails presenting a prompt to the LLM. Here, we adopt the prompt template utilized by~\cite{avalia_varios_prompts}, who, upon evaluating various prompts, concluded that the most effective one contains the task description, examples with respective expected responses, and the text to be evaluated. We adapted and used it for sentiment classification as illustrated in Table~\ref{prompt}. Our prompt consists of instructions and an example for each class (negative and positive) and concludes with the text to be evaluated (Evaluate Text). Subsequently, the LLM generates the class (``next word'') for the evaluated text (Response from LLM).\looseness=-1

\vspace{-0.3cm}
\begin{table}[!h]
\centering
\small
\begin{tabular}{p{7cm}} 
\hline
Classify the sentiment of the following text in the input \newline tag as positive or negative: \newline
<input> I love you.\newline
<output> positive.\newline
<input> The product is bad.\newline
<output> negative. \newline
<input> \{\textbf{Evaluate Text}\}\newline
<output>  \{\textbf{Response from LLM}\}\\
\hline
\end{tabular}
\vspace{-0.2cm}
\caption{Prompt template for sentiment classification.}

\label{prompt}
\end{table}
\vspace{-0.5cm}

\begin{table*}[h!]    
\hspace{-0.3cm}
    \scriptsize
    \centering
        \begin{tabular}{p{1.7cm}|p{1.2cm}|p{1.2cm}|p{1.3cm}|p{1.2cm}|p{1.2cm}|p{1.2cm}|p{1.2cm}|p{1.2cm}}
         \textbf{Dataset} &  \textbf{BART} & \textbf{BERT} & \textbf{RoBERTa} & \textbf{Zero-Shot Llama2} & \textbf{Zero-Shot Bloom} & \textbf{Few-Shot Llama2} & \textbf{Few-Shot Bloom}  & \textbf{Full-Shot (Best)}  \\ 
         \hline
Finance&\textbf{97.0$\pm$1.7}\emp&94.1$\pm$3.8&\textbf{97.8$\pm$1.9}\emp&83.1$\pm$3.2&88.6$\pm$3.1&86.1$\pm$4.3&88.4$\pm$3.3&92.4$\pm$4.2$^{B}$\\     
SemEval2017&\textbf{91.0$\pm$0.4}\emp&90.3$\pm$0.3&\textbf{91.1$\pm$0.5}\emp&79.2$\pm$1.1&78.7$\pm$0.8&82.5$\pm$0.5&81.8$\pm$0.4&\textbf{91.6$\pm$0.5$^{L}$}\\
SiliconeMeldS&\textbf{73.5$\pm$1.7}\emp&\textbf{73.3$\pm$2.1}\emp&72.7$\pm$1.1\emp&58.0$\pm$1.3&69.8$\pm$2.1&59.3$\pm$1.0&70.1$\pm$1.6&\textbf{74.8$\pm$1.8$^{B}$}\\
SiliconeSem&\textbf{85.2$\pm$1.4}\emp&84.7$\pm$1.5&\textbf{86.0$\pm$1.8}\emp&74.6$\pm$2.9&71.6$\pm$1.4&77.5$\pm$4.4&74.3$\pm$2.8&82.1$\pm$1.3$^{L}$\\
SST&87.7$\pm$1.1\emp&86.1$\pm$0.4&87.9$\pm$0.8\emp&77.7$\pm$1.1&88.3$\pm$0.8\emp&79.5$\pm$0.8&88.1$\pm$0.7\emp&\textbf{90.3$\pm$0.7$^{B}$}\\
SST2&94.2$\pm$0.3&\textbf{94.8$\pm$0.1}\emp&\textbf{94.4$\pm$0.3}\emp&80.5$\pm$0.3&89.8$\pm$0.3&85.4$\pm$0.4&90.7$\pm$0.3&93.6$\pm$0.3$^{B}$\\
YelpReview&97.7$\pm$0.2&96.8$\pm$0.4&98.0$\pm$0.4&96.2$\pm$0.9&97.0$\pm$0.9&97.7$\pm$0.4&\textbf{99.0$\pm$0.5}\emp&97.9$\pm$0.2$^{B}$\\
IMDB2024&97.5$\pm$0.6&96.6$\pm$0.5&97.7$\pm$0.6&94.8$\pm$0.5&98.5$\pm$0.6&97.4$\pm$0.6&\textbf{98.9$\pm$0.4}\emp&98.1$\pm$0.6$^{B}$\\
RottenA2023&94.0$\pm$0.7&93.2$\pm$0.5&94.8$\pm$0.8\emp&87.8$\pm$0.8&\textbf{95.4$\pm$0.9}\emp&89.8$\pm$0.9&\textbf{95.6$\pm$0.7}\emp&\textbf{95.8$\pm$0.4$^{B}$}\\
IMDB&92.8$\pm$0.4&91.7$\pm$0.4&93.2$\pm$0.4&88.7$\pm$0.3&93.7$\pm$0.5&92.0$\pm$0.4&\textbf{97.0$\pm$0.3}\emp&96.2$\pm$0.4$^{B}$\\
PangMovie&88.4$\pm$1.0&87.5$\pm$0.7&89.3$\pm$0.6&79.1$\pm$1.3&92.9$\pm$0.6&80.7$\pm$0.8&93.1$\pm$0.7\emp&\textbf{96.0$\pm$0.3$^{B}$}\\

    \end{tabular}
   
    \caption{\footnotesize{Average Macro-F1 and 95\% confidence interval for 1stTRs and open LLMs. Best results (including statistical ties) are marked in \textbf{bold}. \emp marks the best results in a dataset (including statistical ties), removing from the comparison the full-shot (fine-tuning) of the best LLM for each dataset, selected based on the zero-shot performance. The selected LLM in each case is shown with a superscript: $^L$ for LLama2 and $^B$ for Bloom.} }
    \vspace{-0.4cm}\label{transformers}
\end{table*}

\subsection{Tuning Process for Text Classification}

There are basically three different ways of applying pre-trained models, either 1stTRs or LLMs, to text classification tasks: zero-shot, few-shot, and fine-tuning. In the \textbf{zero-shot} strategy, no dataset label is used at any stage, resulting in no adjustment to the model weights. In the \textbf{few-shot} strategy,  a small portion of labeled data is employed to adjust the model weights, simulating a scenario of data scarcity. Lastly, the \textbf{fine-tuning} (a.k.a., full-shot) strategy utilizes all available labeled data in the model's training partition to maximize model adjustment for the task and data domain. While this strategy typically achieves better effectiveness, it has a higher computational cost.\looseness=-1

In the case of 1STRs, we only used them in fine-tuned mode, which is essential for their effectiveness~\cite{artigo_representacao}.  {In this case, fine-tuning consists of learning an appended fully connected head layer (Dense) that captures the distribution of the labels, connecting the ``CLS'' representations with the labels for performing the classification.}\looseness=-1

Specifically for the LLMs, we implemented few-shot using the prompt-based method described above, following the usual literature procedure~\cite{holistic, llama2}, and full-shot LLMs using the same fine-tuning procedure as 1stTRs due to their tendency for higher effectiveness~\cite{tunar_prompt_ou_modelo}.
\looseness=-1

\vspace{-0.2cm}

\subsection{Method-Specific Parameter Tuning}

For 1stTRs, we adopted the hyper-parameterization suggested by ~\cite{cunha23csur}, fixing the learning rate with the value 2$\times10^{-5}$, the batch size with $64$ documents, adjusted the model for five epochs and set the maximum size of each document to 256 tokens. For the LLM models, we adopted the following parameters: for Bloom, we used $2048$ tokens, with a temperature equal to $1$, while for Llama $2$, we used $4096$ maximum tokens with a temperature equal to $0.60$. All other parameters were set at their default values. For few-shot and full-shot processes, which are more costly due to the weight adjustment process of the model (backpropagation), we had to reduce the maximum number of tokens to 256. We performed training for three epochs using the AdaFactor optimizer.\looseness=-1

Our solution has the parameter confidence threshold. Documents for which the 1stTR model does not reach a certain confidence threshold to classify them are sent to the LLM (the Big Sibling). To choose the threshold, we separate part of the training set in a validation set on which we perform classifications varying this parameter to the Macro-F1 metric. The threshold for each dataset can be found in the threshold column in Table~\ref{datasets}. For example, in the SST2 dataset, if the confidence in the prediction is less than 95\%, the document is classified by the LLM. {The higher the threshold, the more documents are sent to the LLM.}\looseness=-1

\vspace{-0.3cm}
\subsection{Metrics and Experimental Protocol}

We evaluated 1stTRs and (open) LLMs regarding the effectiveness/cost tradeoff. Therefore, all models were assessed on identical hardware configuration: a 4-core processor, 32GB of system memory, and an Nvidia Tesla P100 GPU.\looseness=-1

We evaluated the classification effectiveness using Macro Averaged F1 (Macro-F1)~\cite{Sokolova} due to skewness in some datasets. To ensure the statistical validity of the results and demonstrate the generality of the models, we employed a 5-fold stratified cross-validation methodology and the t-test with 95\% confidence.\looseness=-1

To analyze the cost-effectiveness tradeoff, we also evaluated each method's cost in terms of the total time required to build the model. More specifically, the total time comprises the time for model learning (if applicable), together with the time to build the document representation and the time for class prediction (considering the full test set). {Specifically, in the case of our CMBS solution, the time to build the model includes the time to fine-tune the 1stTR, to train the calibrated classifier (which is 1\% of the fine-tuning time, on average), to perform the few-shot with the LLM, if we use this option, and the time to make a prediction.}\looseness=-1

%% file: sections/5_results.tex
\vspace{-0.15cm}
\section{Experimental Results - Analyses}
\label{resultados}
\vspace{-0.15cm}

\subsection{RQ1: Are  (open) LLMs more effective than 1stTRs in sentiment analysis?}

We present the average Macro-F1 results of 1stTRs and the open LLMs in the 11 datasets in Table~\ref{transformers}. Best results (including statistical ties) are marked in \textbf{bold}. Moreover, we adopt~\emp~to mark the best results in a dataset (including statistical ties) for comparison without considering the full-shot results. This last comparison focuses on the results of the 1stTRs, zero and few-shot LLMs. 
The full-shot LLM for each dataset was selected based on the best zero-shot effectiveness result in the respective dataset, and it is  shown with a superscript: $^L$ for Llama2 and $^B$ for Bloom.\looseness=-1

\begin{table*}[h!]   
    \scriptsize
    \centering
        \begin{tabular}{p{1.7cm}|p{1.2cm}|p{1.1cm}|p{1.1cm}|p{1.6cm}|p{1.6cm}|p{1.2cm}|p{1.3cm}|p{1.2cm}}
         \textbf{Dataset} &  \textbf{RoBERTa}  &  \textbf{Zero-Shot Llama2} & \textbf{Zero-Shot Bloom}  
         &\textbf{Few-Shot Llama2}&\textbf{Few-Shot Bloom}&\textbf{Full-Shot (Best)} & \textbf{CMBS (Zero-Shot)} &\textbf{CMBS (Few-Shot)} \\ 
         \hline           Finance&78$\pm$2&54$\pm$0&54$\pm$1&284.3$\pm$9.1&271.4$\pm$7.4&1194$\pm$80&84$\pm$2&282$\pm$5\\
IMDB&2619$\pm$24&4616$\pm$57&3846$\pm$52&6133.7$\pm$505.8&5558.1$\pm$115.3&20107$\pm$64&2847$\pm$27&4948$\pm$87\\
PangMovie&1006$\pm$1&583$\pm$13&1006$\pm$15&1196.5$\pm$78.2&1037.6$\pm$19.6&15367$\pm$1062&1077$\pm$1&1267$\pm$5\\
SemEval2017&2488$\pm$4&1627$\pm$15&1621$\pm$19&2885.3$\pm$98.5&2442.1$\pm$17.8&12766$\pm$76&2672$\pm$4&2845$\pm$9\\
SiliconeMeldS&529$\pm$1&288$\pm$3&559$\pm$17&694.8$\pm$13.9&706.9$\pm$987.7&8078$\pm$530&569$\pm$1&886$\pm$7\\
SiliconeSem&218$\pm$2&116$\pm$0&196$\pm$2&390.7$\pm$32.1&367$\pm$24.7&3414$\pm$9&238$\pm$2&428$\pm$3\\
SST&1031$\pm$1&646$\pm$4&1113$\pm$6&1271.7$\pm$0.7&1124.5$\pm$4&16242$\pm$1124&1106$\pm$1&1297$\pm$9\\
SST2&5830$\pm$14&3224$\pm$19&3004$\pm$27&5972.4$\pm$1354.4&5021.8$\pm$353.3&58944$\pm$135&6261$\pm$14&6416$\pm$15\\
YelpReview&546$\pm$1&572$\pm$5&426$\pm$5&1108.8$\pm$23.7&1020.3$\pm$137.9&6494$\pm$7&591$\pm$1&934$\pm$9\\
IMDB2024&560$\pm$1&928$\pm$1&1151$\pm$5&1537.3$\pm$52.7&1421.2$\pm$58.6&15463$\pm$22&604$\pm$1&932$\pm$11\\
RottenA2023&885$\pm$3&644$\pm$13&568$\pm$9&1163$\pm$70.7&1028$\pm$85.8&10373$\pm$189&963$\pm$4&1132$\pm$12\\
    \end{tabular}
   \vspace{-0.2cm}
    \caption{Average Time total and 95\% confidence interval for RoBERTa,  Zero-Shot Llama2, Zero-Shot Bloom, Few-Shot Llama2, Few-Shot Bloom, Full-Shot best LLM and our solution  CMBS Zero-shot and CMBS Few-shot} 
    \vspace{-0.4cm}
    \label{transformerstempos}
\end{table*}

We can see that, among 1stTRs, RoBERTa achieves the highest effectiveness (or at least ties) in most cases, being the best 1stTR, which is consistent with the literature~\cite{cunha23csur, emlp_roberta_top_sentimento}. Therefore, we will consider RoBERTa as the basis for the comparisons in the next sections, as well as the 1stTR of choice to be used within our CMBS solution.\looseness=-1

Regarding LLMs, we observe that the zero-shot version of both LLMs achieves worse results than full-shot RoBERTa in the vast majority of the cases, with the exception of two ties (SST and RottenA2023 on zero-shot of Bloom). Comparing the two zero-shot LLMs, BLOOM seems to work better for sentiment analysis. This supports its choice for full-shot fine-tuning in 9 out of 11 cases (marked with $^B$ in the last column).\looseness=-1 

Few-shot tuning generates some small improvements in most cases for both LLMs, achieving up to 3.9\% of improvement (in SemEval2017) over zero-shot. These improvements, which come at some cost as we shall discuss next, are not enough to surpass RoBERTa in many cases. The best few-shot LLM loses to RoBERTa in at least 5 datasets - Finance, SemEval2017, SiliconeMeldS and SiliconeSem, SST2, tying in SST.\looseness=-1

Finally, the overall best LLM results (column Full-Shot Best) are obtained when we fine-tune the best zero-shot model between the two alternatives with the full training data. Indeed, there is a general trend for the full-shot to be better than the few-shot -- this happens in 8 out of 11 cases, but three: IMDB, IMDB2024, and YelpReview. In these three datasets, the few-shot version is slightly better than the full-shot one, maybe due to some overfitting or noise in the data. This will be the subject of further investigation in the future.\looseness=-1

Taking up \textbf{RQ1} (\textit{Are (open) LLMs more effective than 1stTRs in sentiment analysis?}), we found that in 8 out of 11 cases some version of the LLMs outperformed or tied with 1stTRs: SemEval2017, SiliconeMeldS, SST, YelpReview, IMDB2024, RottenA2023, IMDB, and PangMovie. Although in our experiments, the  LLMs did not outperform 1stTRs in all datasets, our approach is limited to LLMs that require reasonable structures (in terms of computational resources) for execution (details in Section~\ref{limitation}). Larger models may achieve higher effectiveness~\cite{holistic, llama2} but at higher costs. Indeed, this effectiveness-cost tradeoff is the focus of our analysis in next Section.\looseness=-1

\subsection{RQ2: What is the cost of using open LLMs for ATC in comparison to 1stTR?}
\vspace{-0.1cm}
Table~\ref{transformerstempos} presents the total time (in seconds) for obtaining final predictions for each solution. For now, let´s ignore the last two columns regarding the times for our CMBS solutions. In the Table, we can observe that RoBERTa´s time and the time for running both zero-shots are quite similar, being the shortest ones. LLM Fine-tuning is quite costly, approximately 13 times more expensive compared to fine-tuning RoBERTa. With an average marginal improvement of 0.7\% across all datasets (with a maximum of 7.5\% in PangMovie), it is arguable if such improvements are worth the cost. \looseness=-1  
  
In between (1stTR and full-shot) are the few-shot LLM models, with an average cost (time) around 23\% of RoBERTa's cost, a time similar to the 1stTR in some cases, such as SST2 and PangMovie, and sometimes twice or three times the cost of RoBERTa, such as in Finance and IMDB.\looseness=-1 

The reduced cost of the zero-shot and few-shot compared with the full-shot version of the LLMs, with potential effectiveness gains, especially in the case of few-shot, further motivates our solution.\looseness=-1

\begin{table*}[!h]
\centering
    \includegraphics[width=0.99\textwidth]{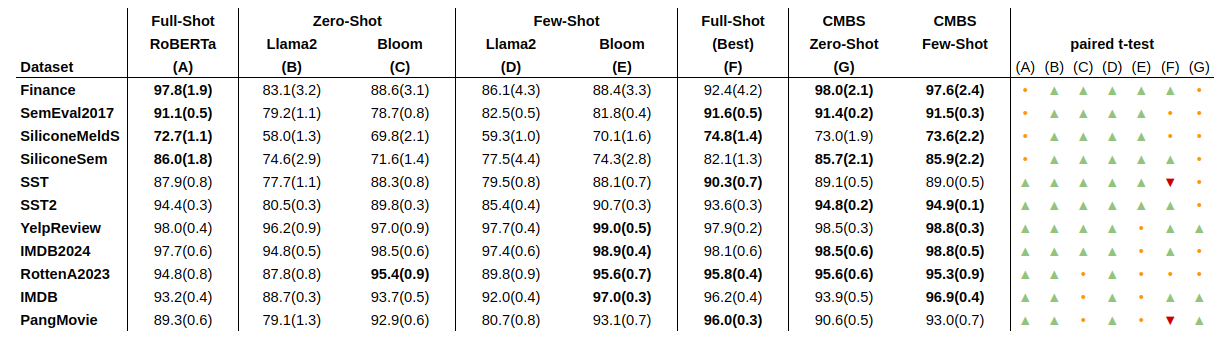}
    \vspace{-0.3cm}
    \caption{\footnotesize{Average Macro-F1 and 95\% confidence interval for first-generation transformers, open LLMs and our proposal (CMBS). Best results (including statistical ties) are marked in \textbf{bold}. In the last column, we employ the paired t-test with 95\% confidence, in which \supe~ means CMBS (Few-Shot) performed better compared to alternative methods (A-G), ~\emp~ means statistical tie, and \infe~ our proposal performed inferior to the alternative methods (A-G).
    \vspace{-0.5cm}
    }}
    \label{nossa_proposta}
\end{table*}

\subsection{RQ3: Is it possible to perform a  1stTRs + LLMs  combination while maintaining a good  effectiveness-cost tradeoff?}

Before delving into the discussion of our CBMS solution, let us analyze the trade-off between the computational cost and the effectiveness of 1sTRs and open LLMs. Figure~\ref{trade_off_modelos} (Appendix~\ref{appendix:tradeoff}) presents  the time in seconds (in decreasing order) on the y-axis and Macro-F1 on the x-axis. Methods that aim to meet both metrics (fast speed and effectiveness) tend to be in the upper right quadrant. We can observe that Full-Shot (black) is by far the most costly solution across all datasets, with moderate effectiveness gains in some datasets (but not all), such as PangMovie, RottenA2023, and SST. {Our most effective solution, the CMBS Few-Shot, is in the upper right quadrant in all cases, demonstrating a good trade-off between effectiveness and efficiency.}  When considering only LLMs,  in scenarios where models do not need to be frequently trained and there is a sufficient computational infrastructure to train such models, fine-tuning is a better solution,  whereas in scenarios with limited computational resources, zero-shot (or few-shot) is a better alternative.\looseness=-1 

Let us focus now on our proposed method: \textbf{CMBS} and compare the two alternative implementations of our solution: CMBS (Zero-Shot) and (Few-Shot), in which low-certainty documents classified by RoBERTa are sent respectively to the best Zero-Shot or Few-Shot LLM in the respective datasets. Macro-F1 results of these alternatives are shown in the last two columns of Table~\ref{nossa_proposta} while the respective costs are shown in the last two columns of Table~\ref{transformerstempos}. As we can see, CMBS (Few-Shot) outperforms its zero-shot counterpart in 3 datasets and ties in 8, with a moderate cost increase of 1.07 on average. Therefore, in the following discussion, we will use CMBS (Few-Shot) in all comparisons as the main representative of our solution.\looseness=-1

We proceed to discuss the results of our solution compared to 1stTRs and LLM Zero-Shot models, which, as previously noted, entail significantly lower computational costs than their tuned versions. In Table~\ref{nossa_proposta}, our solution (statistically) outperforms Zero-Shot LLama2 in all cases in terms of MacroF1, Zero-Shot Bloom in 9 of 11 cases (with two ties) and Roberta in 7 cases (with 4 ties). These very good effectiveness results come with an increase in the cost of only 1.23x over Roberta. Therefore, our solution excels in most cases while maintaining a relatively low cost.\looseness=-1

Figure~\ref{trade_off_modelos} illustrates very well the effectiveness-cost trade-off of our solutions (depicted in green and yellow). As we can see, both CBMS versions are frequently positioned in the upper right quadrant, showing the best balance between the two criteria among all alternatives discussed so far.\looseness=-1

Comparing CMBS with the few-shot version of the LLMs, our solution wins over few-shot Llama 2 in all cases and outperforms the stronger few-shot Bloom in 6 cases, with 5 ties. Regarding cost, the times are usually very close in most datasets, while in others, such as IMDB2024 or IMDB, CMBS (Few-Shot) is up to 1.53 and 1.78 times faster than Few-Shot Bloom, as we send only a few low-certainty documents to the LLM. In this matter, if cost is an issue, CBMS (Zero-Shot) is even faster with an effectiveness close to CBMS (Few-Shot), as we have discussed.\looseness=-1

\begin{figure*}[ht!]
     \centering
     
     \begin{subfigure}[b]{0.28\textwidth}
         \centering
         \includegraphics[width=\textwidth]{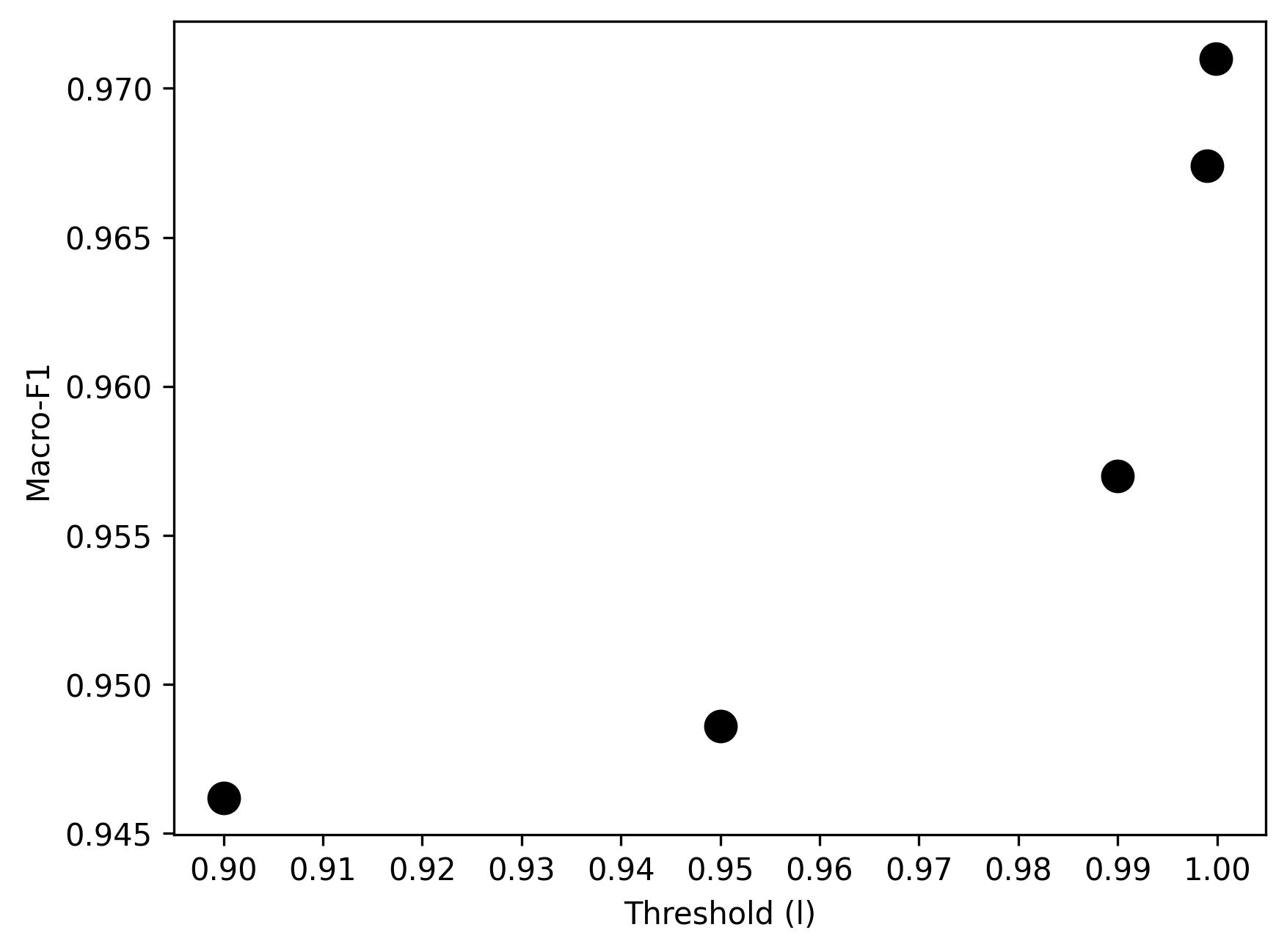}
         \caption{\footnotesize{Effectiveness}}
         \label{fig:efetividade}
     \end{subfigure}
     \begin{subfigure}[b]{0.28\textwidth}
         \centering
         \includegraphics[width=\textwidth]{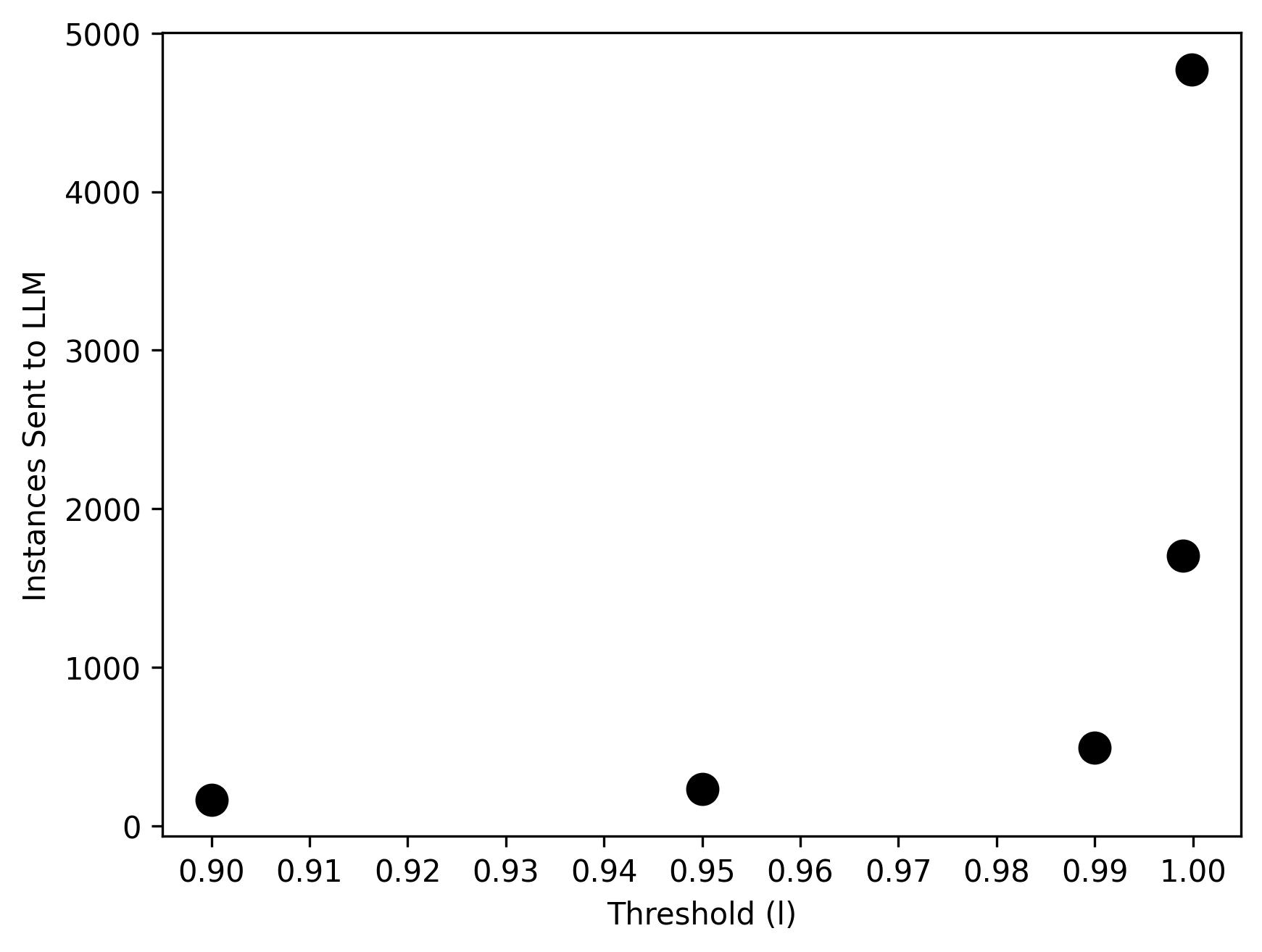}
         \caption{\footnotesize{Instances Sent to LLM}}
         \label{fig:numero_instancias}
     \end{subfigure}     
     \begin{subfigure}[b]{0.28\textwidth}
         \centering
         \includegraphics[width=\textwidth]{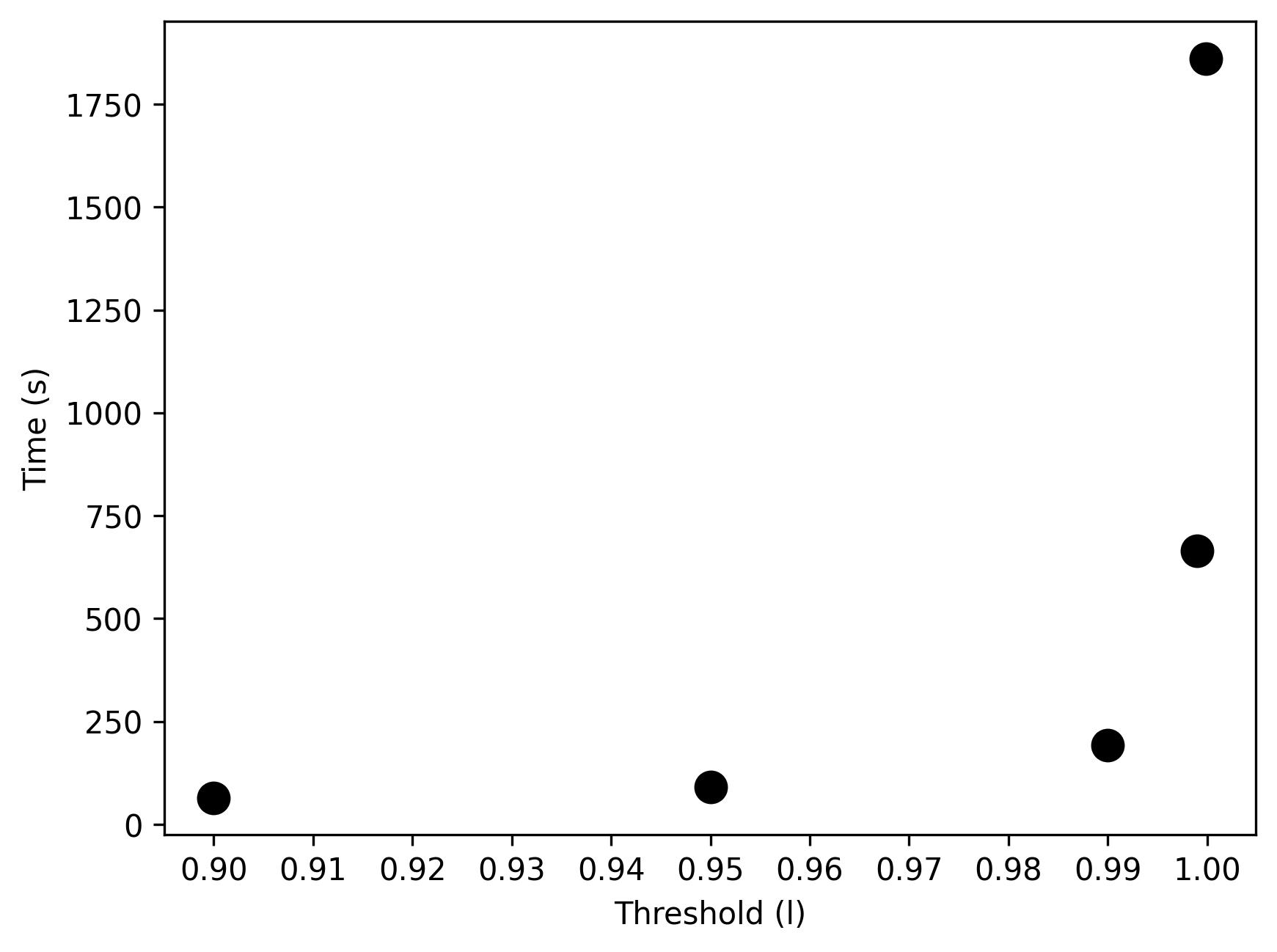}
         \caption{\footnotesize{Efficiency}}
         \label{fig:eficiencia}
     \end{subfigure}     
     \vspace{-0.3cm}
     \caption{Effectiveness, Size of the Test Set Sent to LLM and Efficiency for IMDB dataset.}
     \label{avaliando_limiar}
     \vspace{-0.3cm}
\end{figure*}

To further understand these issues, Table \ref{tab:analysis} shows, per dataset, the number of instances sent to the LLM by CMBS (Few-Shot), the percentage that this number corresponds in the test set,  the Macro-F1 for these instances and the confidence threshold. As we can see in the Table, the number of instances usually sent to the LLM is small, between 2.3\%-30.7\% of the test instances. This is indeed done by design to keep costs low. Only in three datasets, we had to send a larger portion of the test set to the LLM to further improve results -- IMDB, IMDB2024 and PangMovie.  But, even in the last three cases, the costs are significantly lower than the few-shot LLM. In the last column, we see that the Macro-F1 for these low-certainty (hardest) cases is high, being above 67\% in almost all cases (except SE2017 and Sil.Sem) and above 84\% in five cases. If we consider that these are the hardest cases (the ones that RoBERTa is most doubtful about), our results are excellent, which explains the overall improvements with a minimal cost increase.\looseness=-1

\vspace{-0.1cm}
\begin{table}[h!]    
\hspace{-0.4cm}
    \scriptsize
        \begin{tabular}{p{1.7cm}|p{1.1cm}|p{1.3cm}|p{1.3cm}|p{0.8cm}}
         \textbf{Dataset} &    
         \textbf{Instances Sent to LLM} & \textbf{Percentage of Test Instances} & \textbf{LLM Macro-F1} & \textbf{Thre-should (\%)} \\ 
         \hline   
         Finance&4&2.3\%&100.0&95\\
         SemEval2017&239&4.3\%&62.4&95\\
         SiliconeMeldS&364&30.7\%&67.7&95\\ 
         SiliconeSem&59&13.4\%&54.1&95\\
         SST&183&7.7\%&75.3&95\\
         SST2&719&5.3\%&69.7&95\\
         YelpReview&37&3.7\%&89.1&99\\
         IMDB2024&748&56.9\%&91.0&99\\
         RottenA2023&40&2.3\%&77.5&95\\
         IMDB&4280&85.6\%&96.7&99\\         
         PangMovie&1161&54.4\%&92.6&99\\
         
    \end{tabular}
    \vspace{-0.3cm}
    \caption{Instances Sent to LLM by CMBS}
    \vspace{-0.3cm}
    \label{tab:analysis}
\end{table}
\vspace{-0.1cm}

Finally, comparing our solution results with the  Full-Shot Best LLM, we win in three cases, tie in 6 cases and lose in only 2 datasets -- SST and PangMovie at a fraction of the cost $\frac{1}{13}$, as better illustrated in Figure~\ref{trade_off_modelos}. Even in the cases in which we statistically lose to the full-shot, losses are below  1.5\% for SST and 3.3\% for PangMovie. This last dataset, in particular, is one of the few in which the full training fine-tuning really benefited the LLM -- in this case, CBMS ties with few-shot Bloom.\looseness=-1 

In sum, our CBMS solution achieves \textit{``the best of both worlds''}: being as fast as a 1strTR (or close by) while getting very close to the effectiveness of a full-shot LLM without the incurred costs. We also conducted analyses regarding financial costs and carbon emissions of the solutions with similar conclusions (more details in Appendix~\ref{appendix:co2} and \ref{appendix:money}).\looseness=-1

\subsection{Confidence Threshold Sensitivity Analysis}

In this Section, we analyze the role of the uncertainty threshold in the results. We show the results in the IMDB dataset, a dataset in which CBMS (Few-Shot) obtained one of the best cost-benefit tradeoffs: it improves effectiveness over both RoBERTa and Full-Shot (Best) and ties with Few-Shot at almost half the cost of the latter.\looseness=-1

Figure~\ref{fig:efetividade}, \ref{fig:numero_instancias}  and \ref{fig:eficiencia} show the increase in effectiveness, the number of instances sent to LLM, and respective increase in cost. It is very interesting to see that the patterns in increase are very similar in the three graphs, although the metrics are very different. We can also see that by choosing an appropriate threshold, there is still room for improvement, although at the expense of an increase in cost. Finding a perfect balance between these two criteria is not easy and depends on the application requirements. In any case, finding such an optimal point of balance will be the subject of further studies.\looseness=-1

%% file: sections/6_conclusion.tex
\vspace{-0.1cm}
\section{Conclusion}
\label{conclusao}
\vspace{-0.1cm}
We proposed Call-My-Big-Sibling (CMBS), a novel solution that combines the already very effective and efficient 1stTRs with the even more effective, but costlier, open LLMs, aiming at finding a better effectiveness-cost balance in ATC, and specifically in sentiment analysis.    Our approach entails resorting to LLMs only when the utilized 1stTR exhibits uncertainty by relying on calibrated classifiers built on top of 1stTR representations.

Based on experiments conducted on 11 diverse datasets, our findings underscored the superiority of our solution. Notably, we observed seven victories and no defeats compared to the 1stTR (in case, RoBERTa), with a marginal increase in computational time. Moreover, in 9 out of 11 datasets, our method secured the top spot in the overall evaluation when compared to the strongest full-shot LLM at a fraction of the cost- $\frac{1}{13}$ -achieving the ``best of both worlds'' when considering our main goal: a better cost-effectiveness balance.\looseness=-1

In future work, we will apply CMBS to other ATC tasks such as hate speech detection, irony identification, and multi-class topic classification. We will also deepen our investigation into finding the ideal balance between the two considered factors as well as consider other LLMs and alternative ways for performing few-shot with in-context tuning.\looseness=-1

%% file: sections/7_limitation.tex
\newpage
\section{Limitations}
\label{limitation}

Despite relevant contributions, our study has some limitations. Our current work covers only one classification task, which we have pursued to evaluate in depth. In this study, we used 11 datasets belonging to different domains and with distinct characteristics.\looseness=-1

 We focused our evaluation on open LLMs for the sake of the reproducibility of subsequent research using our method. Among LLMs, there are proprietary and closed-source ones, such as ChatGPT, which operate as black boxes. This opacity poses challenges in understanding their training methodologies or internal structures, thereby obstructing reproducibility in research reliant on these models.  \looseness=-1

LLMs have been made available for different purposes. Some of these LLMs have high execution costs, such as Falcon 180B~\cite{falcon}, which requires an expensive infrastructure to use it. In this work, we limited our study to the best evaluated LLMs in the Hugging Face system\footnote{\url{https://huggingface.co/models?pipeline_tag=text-generation&sort=likes}}, with around 7 billion parameters, which have a reasonable structure allowing us to perform zero-shot to full-shot evaluations on our datasets.\looseness=-1

Finally,  our work focused on applying our proposed solution with two open LLMs -- LLama2 and Bloom. However, new LLMs, such as Llama3, emerged during the development of this work, and we were not able to use them in time. We intend to use LLama 3 as well as other new open LLMs that will come out in the near future.  Nevertheless, considering that these new LLMs tend to be increasingly complex and costly, the cost-benefit of our solution will certainly be still valid and even more appealing.\looseness=-1

%% file: sections/8_appendix.tex
\newpage

\appendix

\section{CMBS pseudo-code}
\label{appendix:pseudocode}

The CMBS  pseudo-code is illustrated below.
%
%
%
%
%
%
\vspace{-0.2cm}
	\begin{algorithm}[!h]	
     \small
		\caption{CMBS Algorithm}
		\label{alg:Baseline}
		
		\KwIn{ $D$ (Documents),  $l$ (Sending Limiar Score)}
		\KwOut{ pred (Documents Prediction Set)}

             pred  $\gets \emptyset$\; 
                   
            RoBERTa = RoBERTa.train()\;	
            
		D$_{emb}$ = RoBERTa.representation(D)\; 
  
		proba$_{D}$, predict$_{D}$ = LR(D$_{emb}$)\;
            
		\ForAll{p$_i$ $\in$ proba$_{D}$ }{					
				\If{p$_i$ $<$ $l$}{
					pred$_i$ = LLM.predict(D$_i$) );
				}
                    \Else{
                        pred$_i$ = predict$_{D_i}$;
                    }
		} 
		\Return $pred$;
	\end{algorithm}

\section{Datasets}
\label{appendix:datasets}

Our study draws on \textbf{eleven} datasets developed for binary sentiment classification. Our choice was strategically purposeful due to the effort to perform an in-depth analysis of this task.
The datasets include \textbf{Finance}~\cite{finance} focusing on economic news, \textbf{IMDB}~\cite{IMDB}\footnote{\url{https://www.imdb.com/}} compiling movie reviews as well as \textbf{PangMovie}~\cite{PangMovie} including Rotten Tomatoes\footnote{\url{https://www.rottentomatoes.com/}} data, \textbf{SemEval2017}~\cite{semEval2017} containing Twitter texts used in a significant text classification challenge, \textbf{SiliconeMeldS} and \textbf{SiliconeSem}~\cite{silicone} comprising scripted spoken language (SpokenL)  from films, and the Stanford Sentiment Treebank (\textbf{SST}) \citep{sst2} and \textbf{SST2} \citep{sst2}, where sentiment classification relies on \textit{Treebank},  a corpus with sentiment labels and labeled parse trees.
\textbf{Yelp Review} is a subset of Yelp data widely used in sentiment classification studies \cite{yelp_sergio,yelp_viegas,mendes2020keep}.

\begin{table}[h!]
\hspace{-0.4cm}    
    \scriptsize
    \begin{tabular}{p{1.7cm}|p{1cm}|p{0.7cm}|p{0.7cm}|p{0.62cm}|p{0.62cm}}
        \textbf{Dataset}&\textbf{Domain}&\textbf{|D|}&\textbf{Avg Words}&\textbf{Neg}&\textbf{Pos}\\
        \hline
Finance&Finance&873&24.88&303&570\\
IMDB&Movie&25000&233.78&12500&12500\\
PangMovie&Movie&10662&21.02&5331&5331\\
SemEval2017&Twitter&27743&19.85&7840&19903\\
SiliconeMeldS&SpokenL &5918&11.46&3351&2567\\
SiliconeSem&SpokenL &2197&17.63&936&1261\\
SST&Movie&11855&19.17&5945&5910\\
SST2&Movie&67349&10.41&29780&37569\\
YelpReview &Place&5000&131.7&2500&2500\\
IMDB2024&Movie&6574&163&2057&4517\\
RottenA2023&Movie&8670&49.26&3917&4753\\
    \end{tabular}
    \caption{Datasets Statistics.}
    \label{datasets}
\end{table}

\newpage 

As detailed in Table~\ref{datasets}, we can observe an ample diversity in many aspects in these datasets domain, number of documents (|D|), and density (the average number of words per document).



 \section{CO$_2$ emissions}
 \label{appendix:co2}

 In addition to time, we also calculated the CO$_2$ emissions associated with obtaining final model predictions, using the methodology developed by \citet{lacoste2019quantifying}. Table~\ref{transformers_carbono} presents these values. Similar to time, CO$_2$ emissions are much higher in the fine-tuning of second-generation transformers, in this case, by orders of magnitude. 


\begin{table}[h!]    
    \scriptsize
        \begin{tabular}{l|c|p{1.1cm}|p{1.2cm}|p{1.1cm}}
         \textbf{Dataset} &  \textbf{RoBERTa} & \textbf{Zero-Shot Bloom} & \textbf{Fine-Tunning} & \textbf{CMBS Few-Shot} \\ 
         \hline
         Finance&0.003&0.002&0.046&0.01\\
IMDB&0.101&0.149&0.781&0.121\\
PangMovie&0.039&0.039&0.597&0.049\\
SemEval2017&0.096&0.063&0.496&0.11\\
SiliconeMeldS&0.02&0.021&0.314&0.034\\
SiliconeSem&0.008&0.007&0.132&0.016\\
SST&0.04&0.043&0.631&0.05\\
SST2&0.226&0.116&2.292&0.249\\
YelpReview&0.021&0.016&0.252&0.036\\
IMDB2024&0.021&0.044&0.601&0.036\\
RottenA2023&0.034&0.022&0.403&0.044\\          
    \end{tabular}
    \vspace{-0.3cm}
    \caption{\footnotesize{Emission CO$_2$. Calculation based on the work of \citet{lacoste2019quantifying}}.}
    \label{transformers_carbono}
\end{table}

\begin{table*}[h!]   
    \scriptsize
    \centering
        \begin{tabular}{p{1.7cm}|p{1.2cm}|p{1.1cm}|p{1.1cm}|p{1.1cm}|p{1.1cm}|p{1.2cm}|p{1.3cm}|p{1.2cm}}
         \textbf{Dataset} &  \textbf{RoBERTa}  &  \textbf{Zero-Shot Llama2} & \textbf{Zero-Shot Bloom}  
         &\textbf{Few-Shot Llama2}&\textbf{Few-Shot Bloom}&\textbf{Full-Shot (Best)} & \textbf{CMBS (Zero-Shot)} &\textbf{CMBS (Few-Shot)} \\ 
         \hline     
         Finance&0.08&0.06&0.06&0.3&0.28&1.25&0.09&0.29\\
SemEval2017&2.6&1.7&1.69&3.01&2.55&13.33&2.79&2.97\\
SiliconeMeldS&0.55&0.3&0.58&0.73&0.74&8.44&0.59&0.93\\
SiliconeSem&0.23&0.12&0.2&0.41&0.38&3.57&0.25&0.45\\
SST&1.08&0.67&1.16&1.33&1.17&16.96&1.16&1.35\\
SST2&6.09&3.37&3.14&6.24&5.24&61.56&6.54&6.7\\
YelpReview&0.57&0.6&0.44&1.16&1.07&6.78&0.62&0.98\\
IMDB2024&0.58&0.97&1.2&1.61&1.48&16.15&0.63&0.97\\
RottenA2023&0.92&0.67&0.59&1.21&1.07&10.83&1.01&1.18\\  IMDB&2.74&4.82&4.02&6.41&5.81&21&2.97&3.25\\
PangMovie&1.05&0.61&1.05&1.25&1.08&16.05&1.12&1.32\\  
    \end{tabular}
   
    \caption{Finance Cost in dollars (\$) for RoBERTa,  Zero-Shot Llama2, Zero-Shot Bloom, Few-Shot Llama2, Few-Shot Bloom, Full-Shot best LLM and our solution  CMBS Zero-shot and CMBS Few-shot.} 
    \vspace{-0.4cm}
    \label{custo_financeiro}
\end{table*}

\vspace{-0.4cm}
\section{Finance Cost}
\label{appendix:money}

In the literature, some studies also analyze the financial costs of executing machine learning methods on cloud services~\cite{custo_gpu_aws}. Table~\ref{custo_financeiro} presents the financial cost in dollars for executing the main methods discussed in this paper. We used as a reference the hourly price of a setup similar to the one used in this research~\footnote{\url{https://aws.amazon.com/ec2/instance-types/g4/}}, offered by a large cloud company, which currently costs \$0.752 per hour. The total cost for the main experiments amounted to \$303 (403 hours). As we can observe, the cost of executing only the full-shot method accounts for 58\% (\$176) of the total cost.

\section{Visualizations of the  Trade-offs between effectiveness and computational cost of the 1stTRs  and open LLMs}
~\label{appendix:tradeoff}

\begin{figure*}[ht!]
     \centering
     
     \begin{subfigure}[b]{0.24\textwidth}
         \centering
         \includegraphics[width=\textwidth]{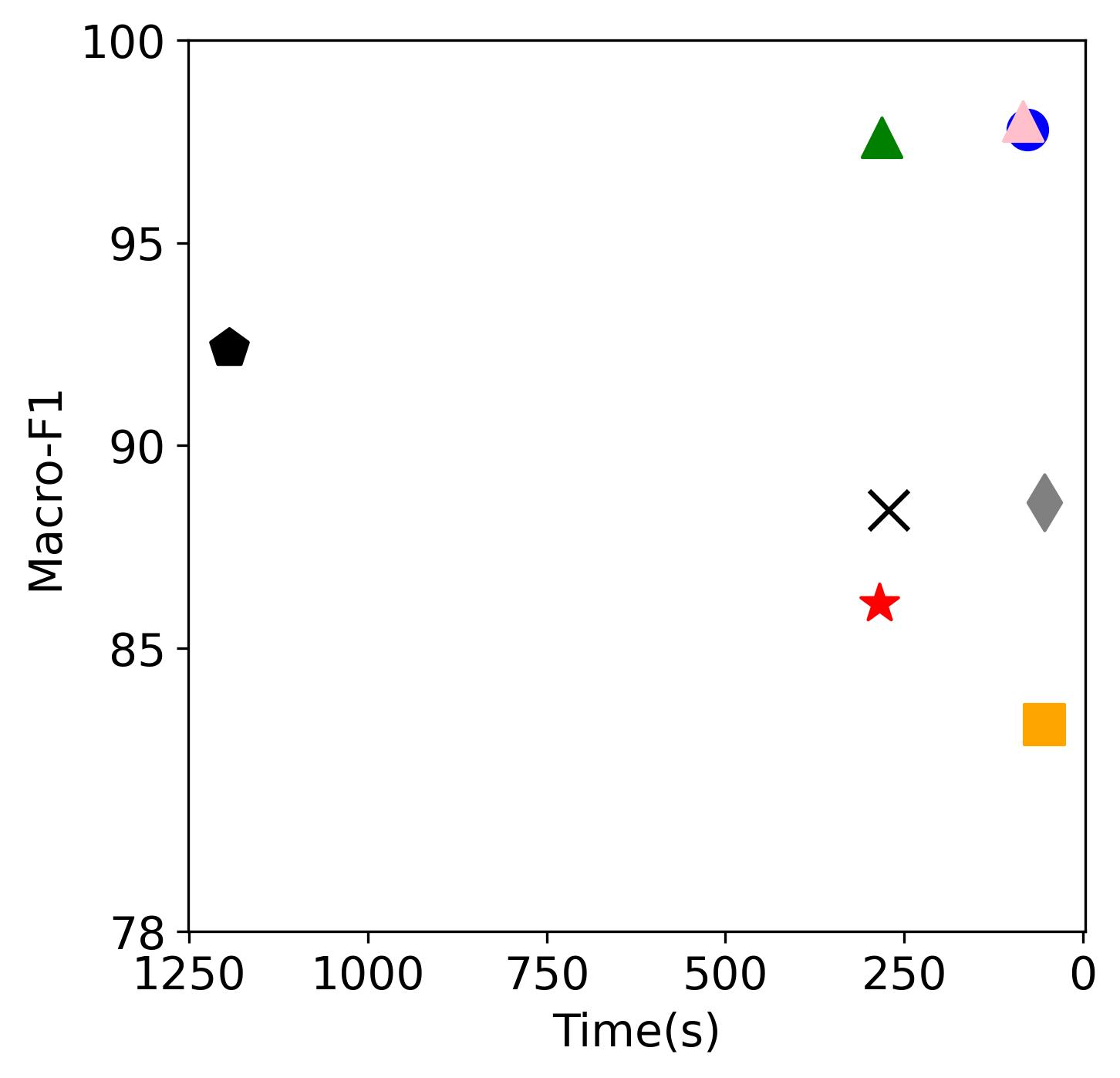}
         \caption{\footnotesize{Finance}}
         \label{fig:y equals x}
     \end{subfigure}
     \begin{subfigure}[b]{0.24\textwidth}
         \centering
         \includegraphics[width=\textwidth]{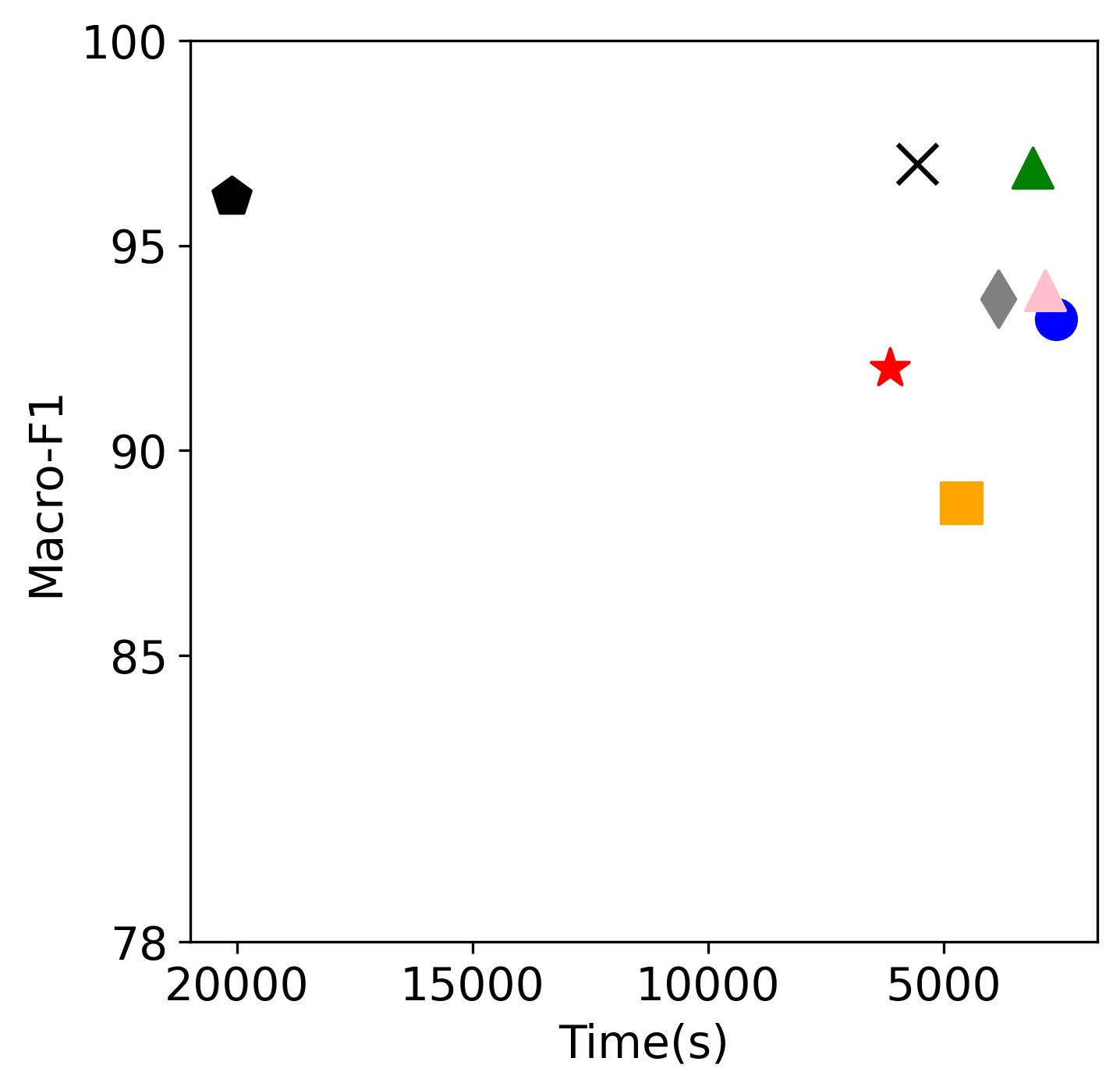}
         \caption{\footnotesize{IMDB}}
         \label{fig:three sin x}
     \end{subfigure}     
     \begin{subfigure}[b]{0.24\textwidth}
         \centering
         \includegraphics[width=\textwidth]{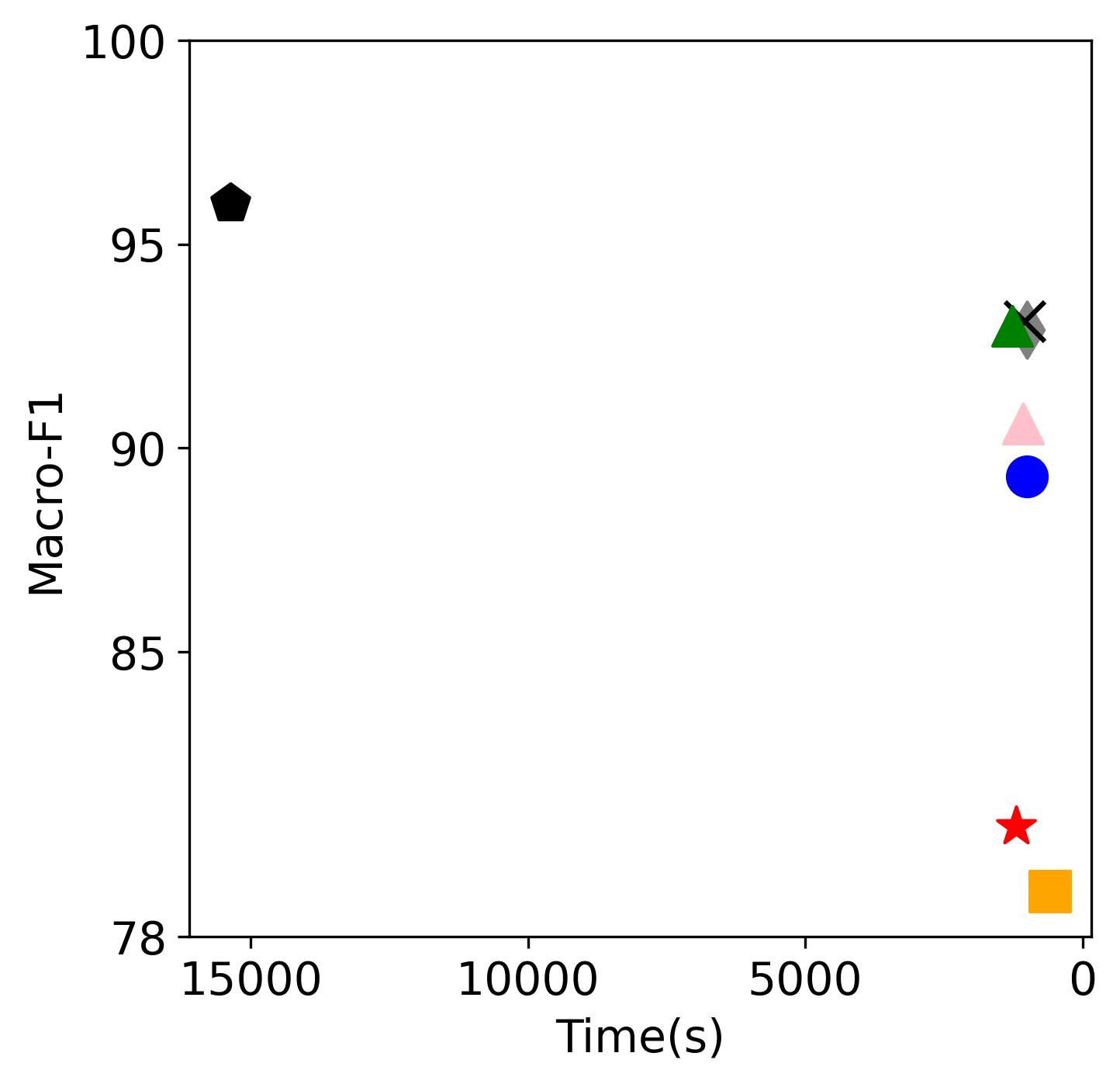}
         \caption{\footnotesize{Pang Movie}}
         \label{fig:three sin x}
     \end{subfigure}
     \begin{subfigure}[b]{0.24\textwidth}
         \centering
         \includegraphics[width=\textwidth]{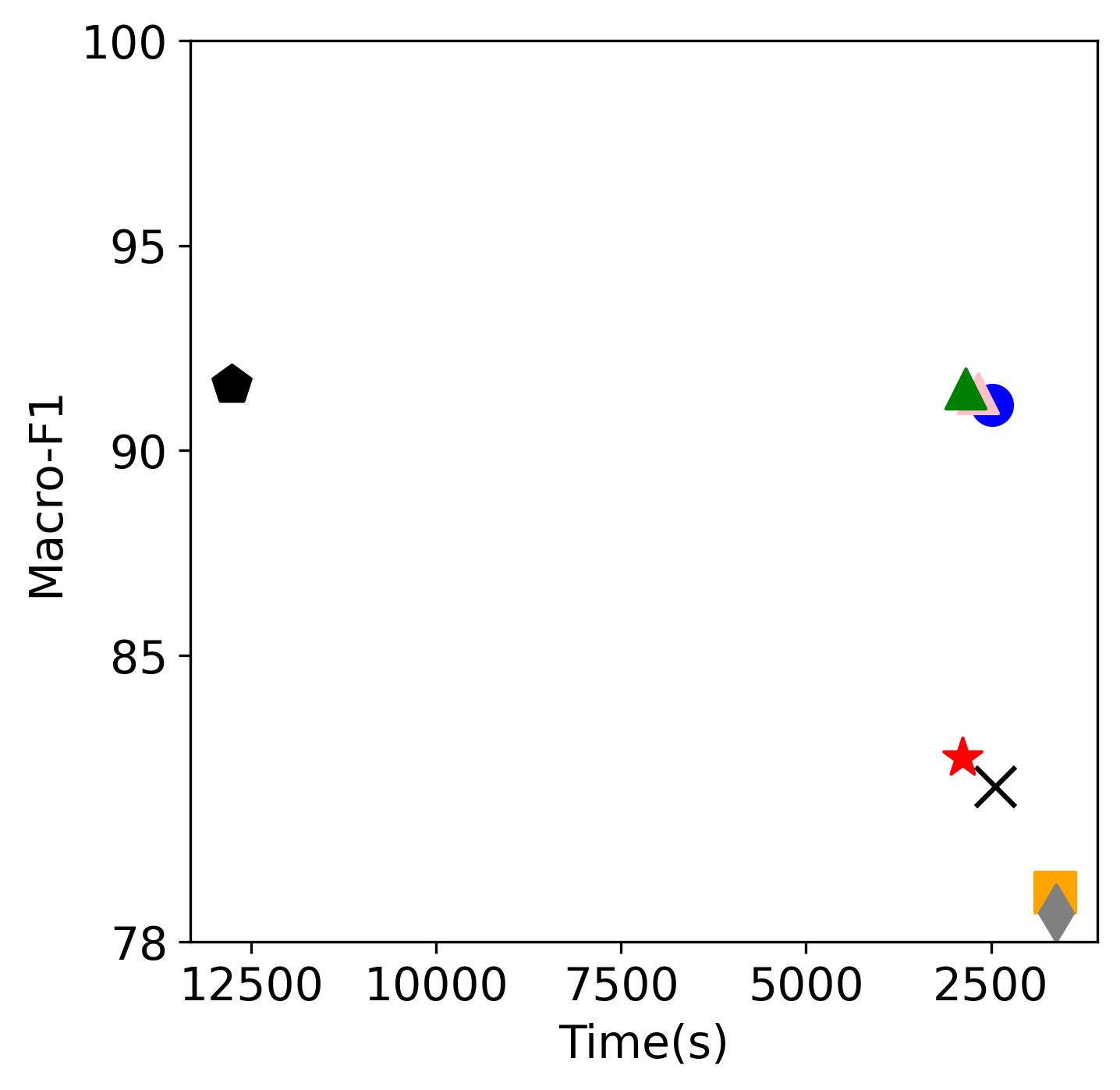}
         \caption{\footnotesize{SemEval 2017}}
         \label{fig:three sin x}
     \end{subfigure}
      \begin{subfigure}[b]{0.24\textwidth}
         \centering
         \includegraphics[width=\textwidth]{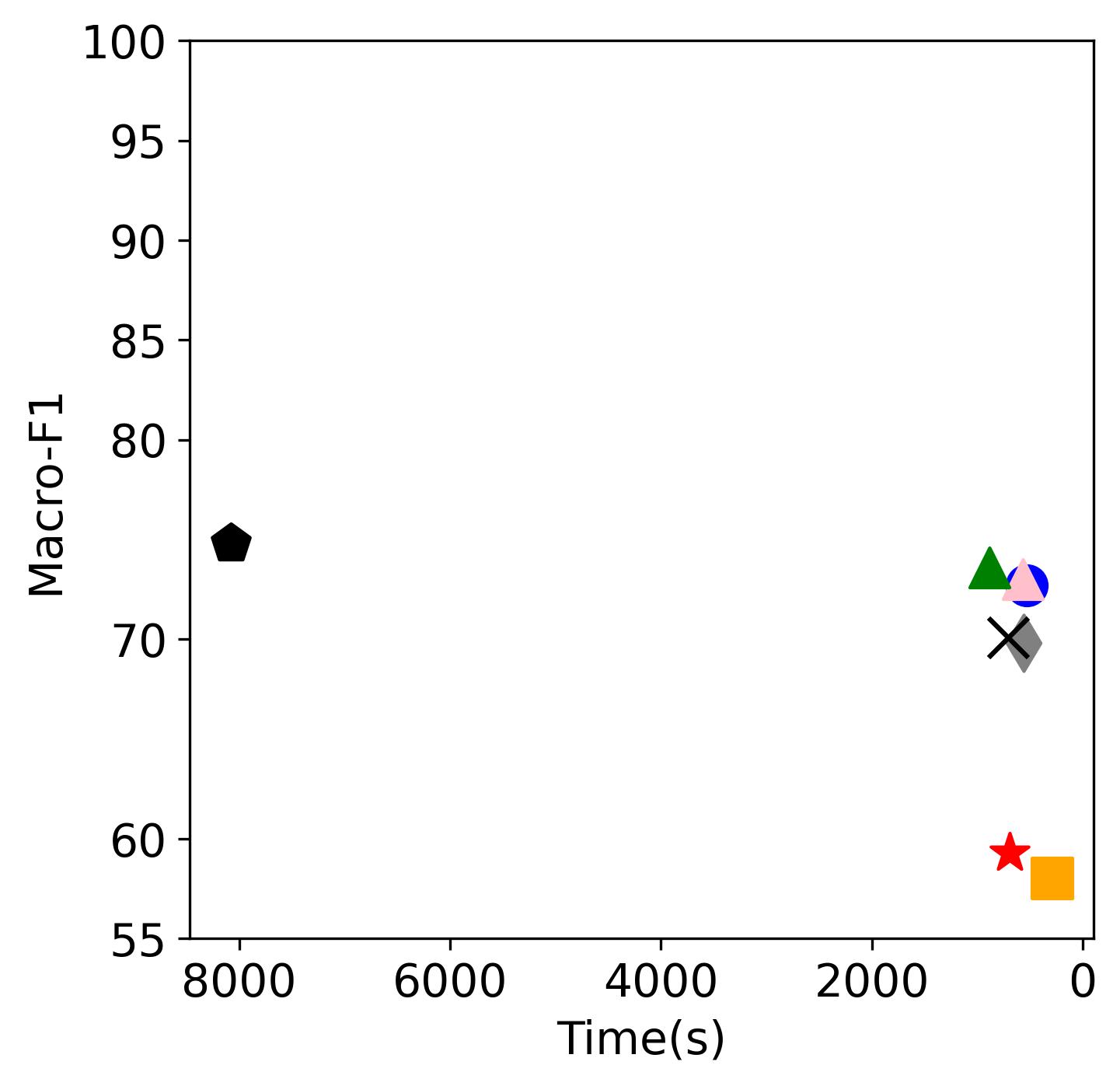}
         \caption{\footnotesize{SiliconeMeldS}}
         \label{fig:three sin x}
     \end{subfigure}
      \begin{subfigure}[b]{0.24\textwidth}
         \centering
         \includegraphics[width=\textwidth]{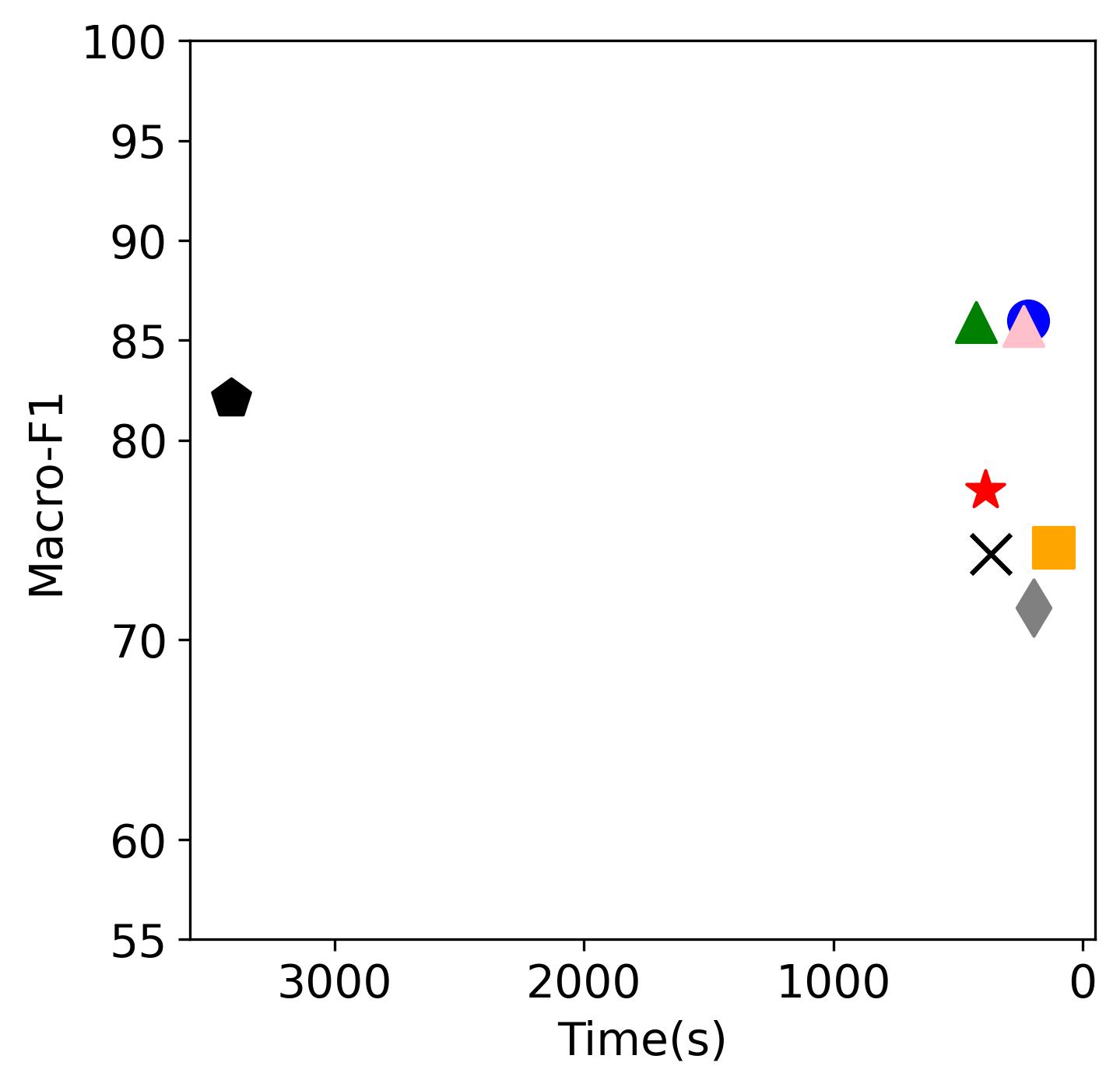}
         \caption{\footnotesize{SiliconeSem}}
         \label{fig:three sin x}
     \end{subfigure}
      \begin{subfigure}[b]{0.24\textwidth}
         \centering
         \includegraphics[width=\textwidth]{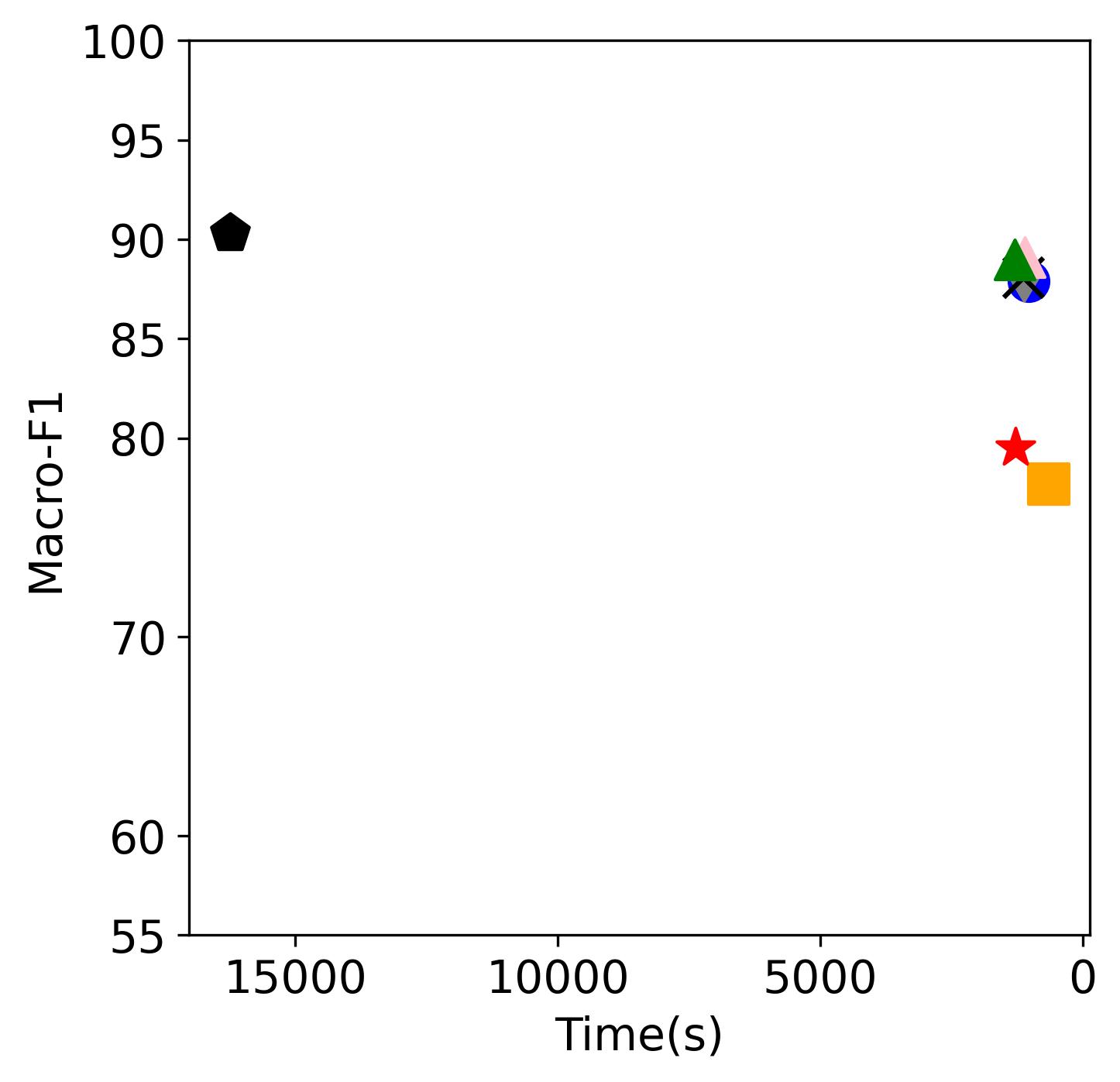}
         \caption{\footnotesize{SST}}
         \label{fig:three sin x}
     \end{subfigure}
      \begin{subfigure}[b]{0.24\textwidth}
         \centering
         \includegraphics[width=\textwidth]{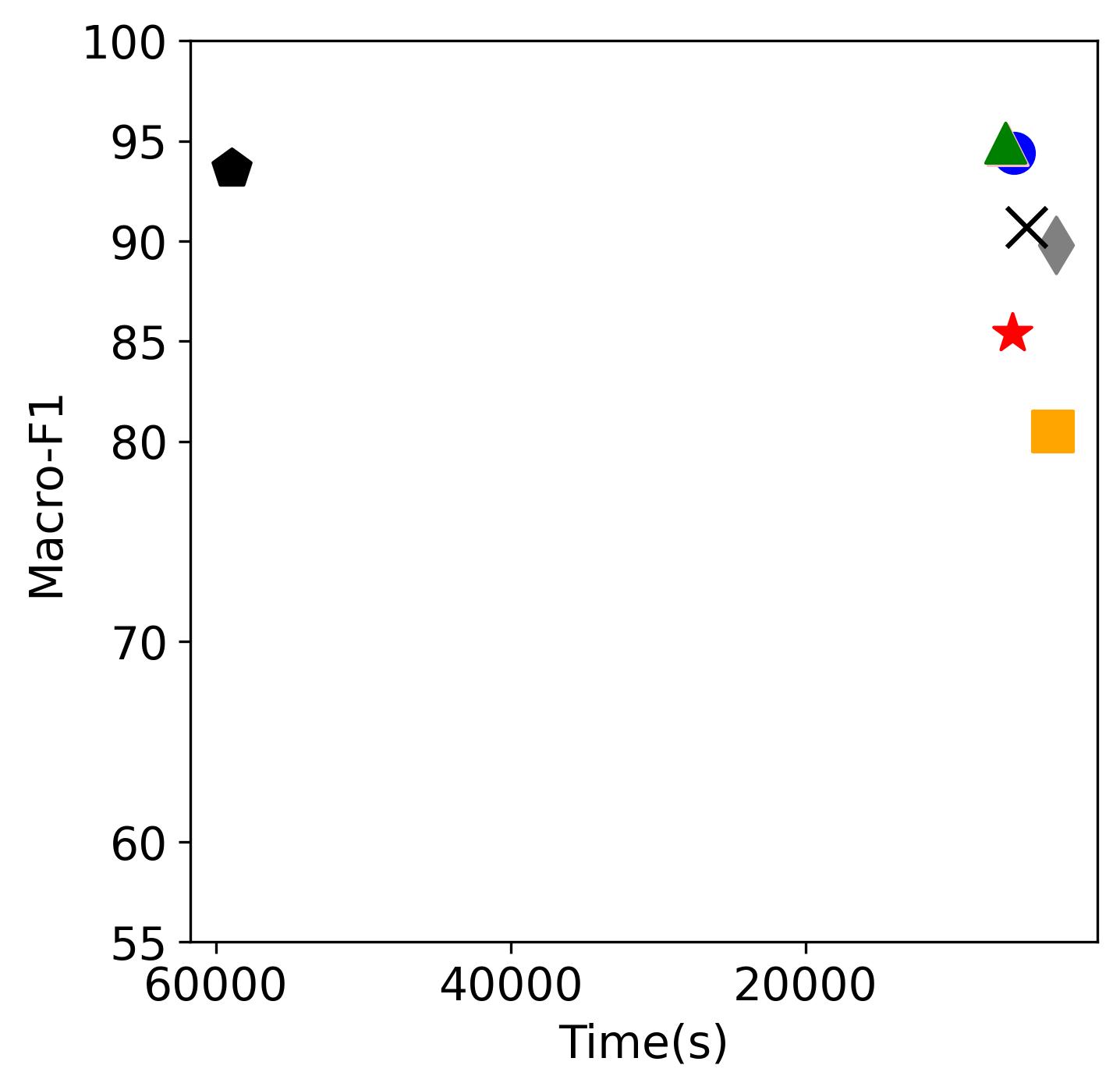}
         \caption{\footnotesize{SST2}}
         \label{fig:three sin x}
     \end{subfigure}      
      \begin{subfigure}[b]{0.24\textwidth}
         \centering
         \includegraphics[width=\textwidth]{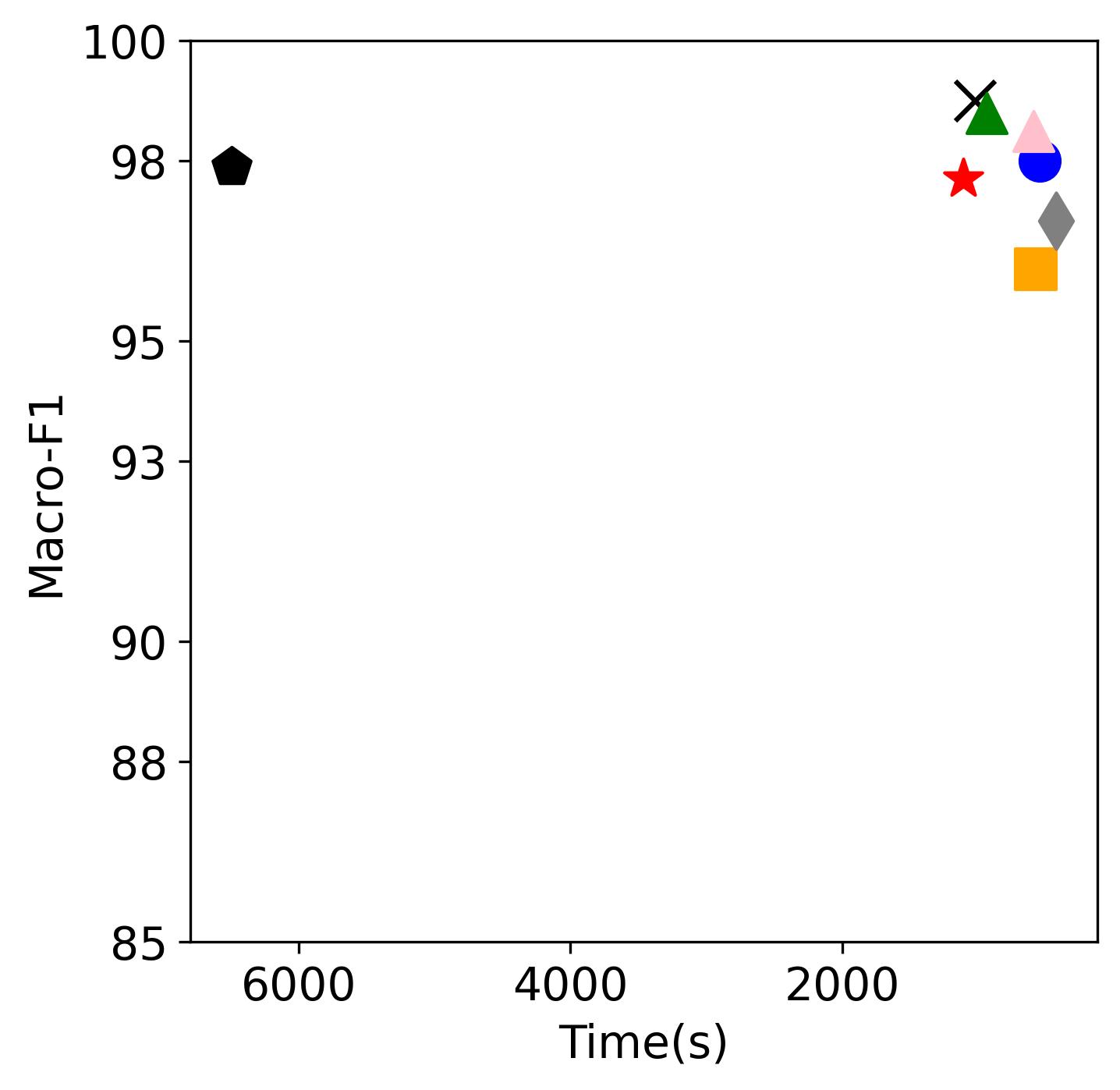}
         \caption{\footnotesize{YelpReview}}
         \label{fig:three sin x}
     \end{subfigure}
      \begin{subfigure}[b]{0.24\textwidth}
         \centering
         \includegraphics[width=\textwidth]{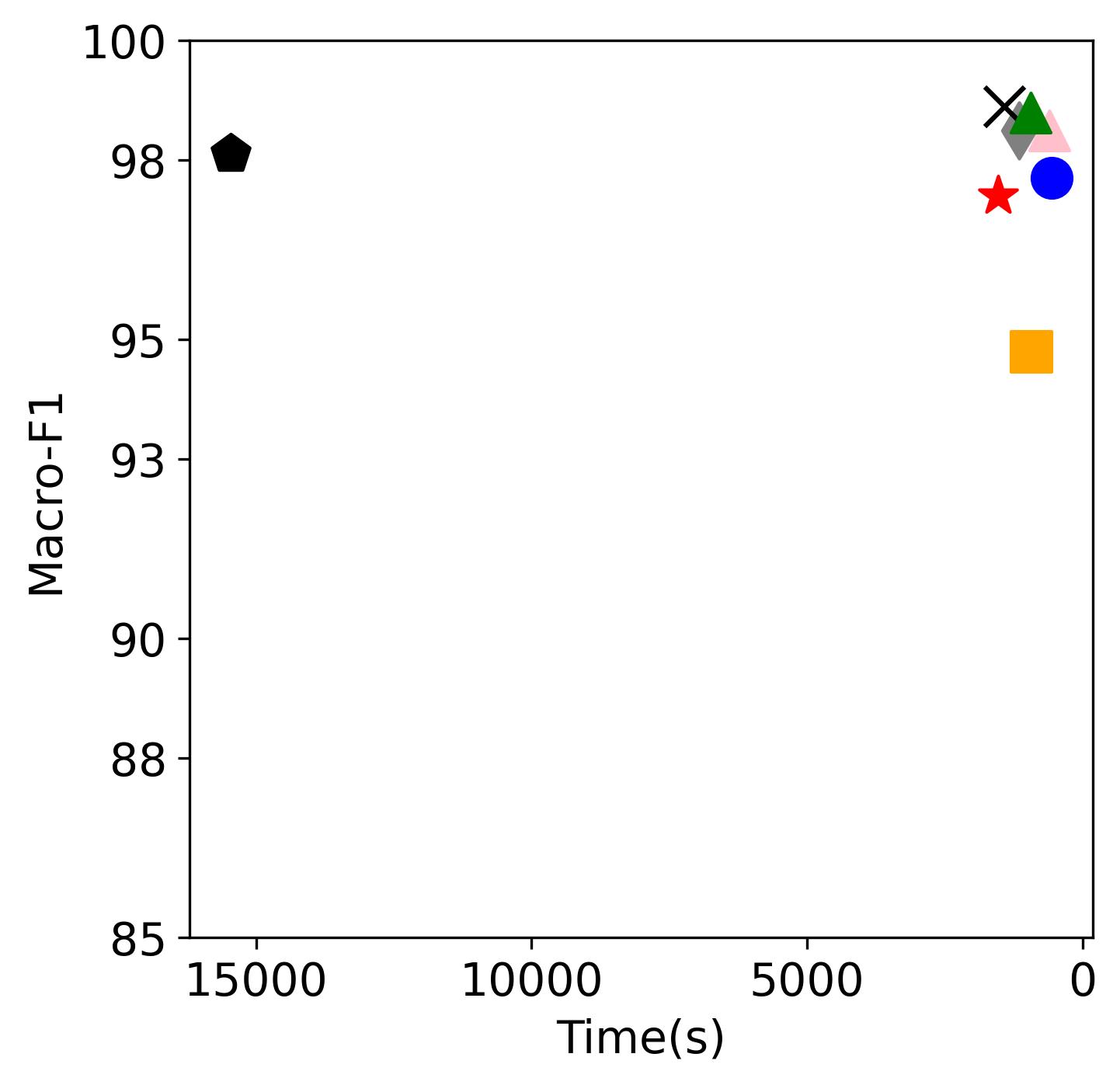}
         \caption{\footnotesize{IMDB2024}}
         \label{fig:three sin x}
     \end{subfigure}
      \begin{subfigure}[b]{0.24\textwidth}
         \centering
         \includegraphics[width=\textwidth]{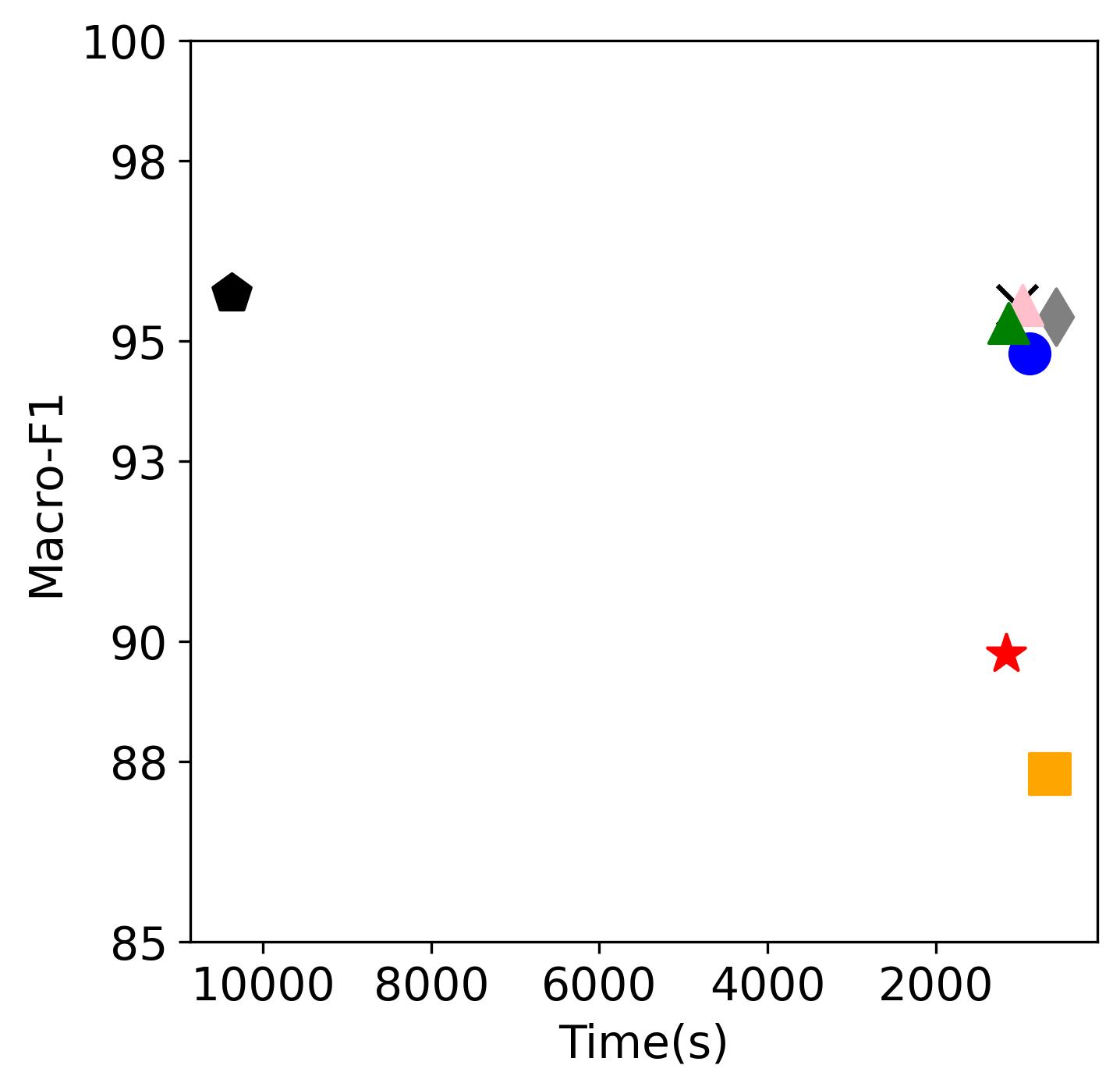}
         \caption{\footnotesize{RottenA2023}}
         \label{fig:three sin x}
     \end{subfigure}
     \begin{subfigure}[b]{0.85\textwidth}
         \centering
         \includegraphics[width=\textwidth]{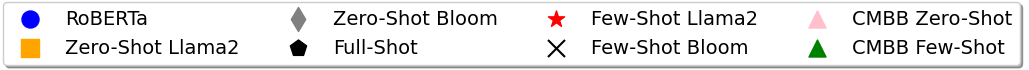}
     \end{subfigure} 
     \caption{Total Time (seconds) and Macro-F1 in RoBERTa, Zero-Shot Llama2, Zero-Shot Bloom, Fine-tunning LLM and our  proposed solution.}
     \label{trade_off_modelos}
     \vspace{-0.2cm}
\end{figure*}

\noindent Figure~\ref{trade_off_modelos} presents time in seconds (in decreasing order) on the y-axis and MacroF1 on the x-axis. In this graph, methods that aim to meet both metrics (fast and effective) tend to be in the upper right quadrant.

  \vspace{-0.4cm}

%% file: main.bbl
\begin{thebibliography}{41}
\expandafter\ifx\csname natexlab\endcsname\relax\def\natexlab#1{#1}\fi

\bibitem[{Bai et~al.(2023{\natexlab{a}})Bai, Zhang, Li, Hong, Xu, Lin, and
  Rong}]{roberta_forte}
Jun Bai, Xiaofeng Zhang, Chen Li, Hanhua Hong, Xi~Xu, Chenghua Lin, and Wenge
  Rong. 2023{\natexlab{a}}.
\newblock \href {https://doi.org/10.18653/v1/2023.findings-emnlp.357} {How to
  determine the most powerful pre-trained language model without brute force
  fine-tuning? an empirical survey}.
\newblock In \emph{Findings of the Association for Computational Linguistics:
  EMNLP 2023}, pages 5369--5382, Singapore. Association for Computational
  Linguistics.

\bibitem[{Bai et~al.(2023{\natexlab{b}})Bai, Zhang, Li, Hong, Xu, Lin, and
  Rong}]{emlp_roberta_top_sentimento}
Jun Bai, Xiaofeng Zhang, Chen Li, Hanhua Hong, Xi~Xu, Chenghua Lin, and Wenge
  Rong. 2023{\natexlab{b}}.
\newblock \href {https://doi.org/10.18653/v1/2023.findings-emnlp.357} {How to
  determine the most powerful pre-trained language model without brute force
  fine-tuning? an empirical survey}.
\newblock In \emph{Findings of the EMNLP 2023}.

\bibitem[{Bel{\'e}m et~al.(2024)Bel{\'e}m, Cunha, Fran{\c{c}}a, Andrade, Rocha,
  and Gon{\c{c}}alves}]{belem2024novel}
Fabiano Bel{\'e}m, Washington Cunha, Celso Fran{\c{c}}a, Claudio Andrade,
  Leonardo Rocha, and Marcos~Andr{\'e} Gon{\c{c}}alves. 2024.
\newblock A novel two-step fine-tuning pipeline for cold-start active learning
  in text classification tasks.
\newblock \emph{arXiv preprint arXiv:2407.17284}.

\bibitem[{Biswas et~al.(2023)Biswas, Islam, Shah, Zaghouani, and
  Belhaouari}]{biswas2023can}
Md~Rafiul Biswas, Ashhadul Islam, Zubair Shah, Wajdi Zaghouani, and
  Samir~Brahim Belhaouari. 2023.
\newblock Can chatgpt be your personal medical assistant?
\newblock In \emph{2023 Tenth International Conference on Social Networks
  Analysis, Management and Security (SNAMS)}, pages 1--5. IEEE.

\bibitem[{Canuto et~al.(2016)Canuto, Gon{\c{c}}alves, and
  Benevenuto}]{yelp_sergio}
S{\'{e}}rgio~D. Canuto, Marcos~Andr{\'{e}} Gon{\c{c}}alves, and Fabr{\'{\i}}cio
  Benevenuto. 2016.
\newblock \href {https://doi.org/10.1145/2835776.2835821} {Exploiting new
  sentiment-based meta-level features for effective sentiment analysis}.
\newblock In \emph{Proceedings of the Ninth {ACM} International Conference on
  Web Search and Data Mining, San Francisco, CA, USA, February 22-25, 2016},
  pages 53--62. {ACM}.

\bibitem[{Chapuis et~al.(2020)Chapuis, Colombo, Manica, Labeau, and
  Clavel}]{silicone}
Emile Chapuis, Pierre Colombo, Matteo Manica, Matthieu Labeau, and Chlo{\'{e}}
  Clavel. 2020.
\newblock \href {https://doi.org/10.18653/V1/2020.FINDINGS-EMNLP.239}
  {Hierarchical pre-training for sequence labelling in spoken dialog}.
\newblock In \emph{Findings of the Association for Computational Linguistics:
  {EMNLP} 2020, Online Event, 16-20 November 2020}, volume {EMNLP} 2020 of
  \emph{Findings of {ACL}}, pages 2636--2648. Association for Computational
  Linguistics.

\bibitem[{Chen et~al.(2021)Chen, Dong, Qiu, He, Xin, Chen, Lin, and
  Yang}]{10.1145/3404835.3462919}
Jiawei Chen, Hande Dong, Yang Qiu, Xiangnan He, Xin Xin, Liang Chen, Guli Lin,
  and Keping Yang. 2021.
\newblock Autodebias: Learning to debias for recommendation.
\newblock In \emph{Proceedings of the 44th International ACM SIGIR Conference
  on Research and Development in Information Retrieval}, SIGIR '21, page
  21–30, New York, NY, USA. Association for Computing Machinery.

\bibitem[{Chen and Li(2023)}]{WenlongICLR2024}
Wenlong Chen and Yingzhen Li. 2023.
\newblock Calibrating transformers via sparse gaussian processes.
\newblock \emph{arXiv preprint arXiv:2303.02444}.

\bibitem[{Cunha et~al.(2020)Cunha, Canuto, Viegas, Salles, Gomes, Mangaravite,
  Resende, Rosa, Gon{\c{c}}alves, and Rocha}]{cunha2020extended}
Washington Cunha, S{\'e}rgio Canuto, Felipe Viegas, Thiago Salles, Christian
  Gomes, Vitor Mangaravite, Elaine Resende, Thierson Rosa, Marcos~Andr{\'e}
  Gon{\c{c}}alves, and Leonardo Rocha. 2020.
\newblock Extended pre-processing pipeline for text classification: On the role
  of meta-feature representations, sparsification and selective sampling.
\newblock \emph{Information Processing \& Management}, 57(4):102263.

\bibitem[{Cunha et~al.(2023{\natexlab{a}})Cunha, Fran{\c{c}}a, Fonseca, Rocha,
  and Gon{\c{c}}alves}]{cunha2023effective}
Washington Cunha, Celso Fran{\c{c}}a, Guilherme Fonseca, Leonardo Rocha, and
  Marcos~Andr{\'e} Gon{\c{c}}alves. 2023{\natexlab{a}}.
\newblock An effective, efficient, and scalable confidence-based instance
  selection framework for transformer-based text classification.
\newblock In \emph{Proceedings of the 46th International ACM SIGIR Conference
  on Research and Development in Information Retrieval}, pages 665--674.

\bibitem[{Cunha et~al.(2023{\natexlab{b}})Cunha, Fran{\c{c}}a, Rocha, and
  Gon{\c{c}}alves}]{cunha2023tpdr}
Washington Cunha, Celso Fran{\c{c}}a, Leonardo Rocha, and Marcos~Andr{\'e}
  Gon{\c{c}}alves. 2023{\natexlab{b}}.
\newblock Tpdr: A novel two-step transformer-based product and class
  description match and retrieval method.
\newblock \emph{arXiv preprint arXiv:2310.03491}.

\bibitem[{Cunha et~al.(2021)Cunha, Mangaravite, Gomes, Canuto, Resende,
  Nascimento, Viegas, Fran{\c{c}}a, Martins, Almeida et~al.}]{cunha2021cost}
Washington Cunha, V{\'\i}tor Mangaravite, Christian Gomes, S{\'e}rgio Canuto,
  Elaine Resende, Cecilia Nascimento, Felipe Viegas, Celso Fran{\c{c}}a,
  Wellington~Santos Martins, Jussara~M Almeida, et~al. 2021.
\newblock On the cost-effectiveness of neural and non-neural approaches and
  representations for text classification: A comprehensive comparative study.
\newblock \emph{Information Processing \& Management}, 58(3):102481.

\bibitem[{Cunha et~al.(2023{\natexlab{c}})Cunha, Viegas, Fran\c{c}a, Rosa,
  Rocha, and Gon\c{c}alves}]{cunha23csur}
Washington Cunha, Felipe Viegas, Celso Fran\c{c}a, Thierson Rosa, Leonardo
  Rocha, and Marcos~Andr\'{e} Gon\c{c}alves. 2023{\natexlab{c}}.
\newblock \href {https://doi.org/10.1145/3582000} {A comparative survey of
  instance selection methods applied to non-neural and transformer-based text
  classification}.
\newblock \emph{ACM CSUR}.

\bibitem[{de~Andrade and Gon{\c{c}}alves(2021)}]{Andrade21}
Claudio Mois{\'{e}}s~Valiense de~Andrade and Marcos~Andr{\'{e}}
  Gon{\c{c}}alves. 2021.
\newblock \href {http://ceur-ws.org/Vol-2936/paper-195.pdf} {Profiling hate
  speech spreaders on twitter: Exploiting textual analysis of tweets and
  combination of textual representations}.
\newblock In \emph{Proceedings of the Working Notes of {CLEF} 2021 - Conference
  and Labs of the Evaluation Forum, Bucharest, Romania, September 21st - to -
  24th, 2021}, volume 2936 of \emph{{CEUR} Workshop Proceedings}, pages
  2186--2192. CEUR-WS.org.

\bibitem[{{de Andrade} et~al.(2023){de Andrade}, Belém, Cunha, França,
  Viegas, Rocha, and Gonçalves}]{artigo_representacao}
Claudio~M.V. {de Andrade}, Fabiano~M. Belém, Washington Cunha, Celso França,
  Felipe Viegas, Leonardo Rocha, and Marcos~André Gonçalves. 2023.
\newblock \href {https://doi.org/https://doi.org/10.1016/j.ipm.2023.103336} {On
  the class separability of contextual embeddings representations – or “the
  classifier does not matter when the (text) representation is so good!”}.
\newblock \emph{Information Processing \& Management}, 60(4):103336.

\bibitem[{Devlin et~al.(2019)Devlin, Chang, Lee, and Toutanova}]{Devlin_2018}
Jacob Devlin, Ming-Wei Chang, Kenton Lee, and Kristina Toutanova. 2019.
\newblock \href {https://doi.org/10.18653/v1/N19-1423} {{BERT}: Pre-training of
  deep bidirectional transformers for language understanding}.
\newblock pages 4171--4186.

\bibitem[{Fields et~al.(2024)Fields, Chovanec, and Madiraju}]{survey}
John Fields, Kevin Chovanec, and Praveen Madiraju. 2024.
\newblock \href {https://doi.org/10.1109/ACCESS.2024.3349952} {A survey of text
  classification with transformers: How wide? how large? how long? how
  accurate? how expensive? how safe?}
\newblock \emph{{IEEE} Access}, 12:6518--6531.

\bibitem[{Fran{\c{c}}a et~al.(2024)Fran{\c{c}}a, Lima, Andrade, Cunha, de~Melo,
  Ribeiro-Neto, Rocha, Santos, Pagano, and
  Gon{\c{c}}alves}]{francca2024representation}
Celso Fran{\c{c}}a, Rennan~C Lima, Claudio Andrade, Washington Cunha, Pedro
  OS~Vaz de~Melo, Berthier Ribeiro-Neto, Leonardo Rocha, Rodrygo~LT Santos,
  Adriana~Silvina Pagano, and Marcos~Andr{\'e} Gon{\c{c}}alves. 2024.
\newblock On representation learning-based methods for effective, efficient,
  and scalable code retrieval.
\newblock \emph{Neurocomputing}, 600:128172.

\bibitem[{Gao et~al.(2024)Gao, Hu, Ruan, Pu, and Wan}]{gao2024llm}
Mingqi Gao, Xinyu Hu, Jie Ruan, Xiao Pu, and Xiaojun Wan. 2024.
\newblock Llm-based nlg evaluation: Current status and challenges.
\newblock \emph{arXiv preprint arXiv:2402.01383}.

\bibitem[{Griggs et~al.(2024)Griggs, Liu, Yu, Kim, Chiang, Cheung, and
  Stoica}]{custo_gpu_aws}
Tyler Griggs, Xiaoxuan Liu, Jiaxiang Yu, Doyoung Kim, Wei{-}Lin Chiang, Alvin
  Cheung, and Ion Stoica. 2024.
\newblock \href {https://doi.org/10.48550/ARXIV.2404.14527}
  {M{\textbackslash}'elange: Cost efficient large language model serving by
  exploiting {GPU} heterogeneity}.
\newblock \emph{CoRR}, abs/2404.14527.

\bibitem[{Hajba and Hajba(2018)}]{hajba2018using}
G{\'a}bor~L{\'a}szl{\'o} Hajba and G{\'a}bor~L{\'a}szl{\'o} Hajba. 2018.
\newblock Using beautiful soup.
\newblock \emph{Website Scraping With Python: Using BeautifulSoup and Scrapy},
  pages 41--96.

\bibitem[{Lacoste et~al.(2019)Lacoste, Luccioni, Schmidt, and
  Dandres}]{lacoste2019quantifying}
Alexandre Lacoste, Alexandra Luccioni, Victor Schmidt, and Thomas Dandres.
  2019.
\newblock Quantifying the carbon emissions of machine learning.
\newblock \emph{arXiv preprint arXiv:1910.09700}.

\bibitem[{Lester et~al.(2021)Lester, Al{-}Rfou, and
  Constant}]{tunar_prompt_ou_modelo}
Brian Lester, Rami Al{-}Rfou, and Noah Constant. 2021.
\newblock \href {http://arxiv.org/abs/2104.08691} {The power of scale for
  parameter-efficient prompt tuning}.
\newblock \emph{CoRR}, abs/2104.08691.

\bibitem[{Liang et~al.(2023)Liang, Bommasani, Lee, Tsipras, Soylu, Yasunaga,
  Zhang, Narayanan, Wu, Kumar, Newman, Yuan, Yan, Zhang, Cosgrove, Manning, Re,
  Acosta-Navas, Hudson, Zelikman, Durmus, Ladhak, Rong, Ren, Yao, WANG,
  Santhanam, Orr, Zheng, Yuksekgonul, Suzgun, Kim, Guha, Chatterji, Khattab,
  Henderson, Huang, Chi, Xie, Santurkar, Ganguli, Hashimoto, Icard, Zhang,
  Chaudhary, Wang, Li, Mai, Zhang, and Koreeda}]{holistic}
Percy Liang, Rishi Bommasani, Tony Lee, Dimitris Tsipras, Dilara Soylu,
  Michihiro Yasunaga, Yian Zhang, Deepak Narayanan, Yuhuai Wu, Ananya Kumar,
  Benjamin Newman, Binhang Yuan, Bobby Yan, Ce~Zhang, Christian~Alexander
  Cosgrove, Christopher~D Manning, Christopher Re, Diana Acosta-Navas,
  Drew~Arad Hudson, Eric Zelikman, Esin Durmus, Faisal Ladhak, Frieda Rong,
  Hongyu Ren, Huaxiu Yao, Jue WANG, Keshav Santhanam, Laurel Orr, Lucia Zheng,
  Mert Yuksekgonul, Mirac Suzgun, Nathan Kim, Neel Guha, Niladri~S. Chatterji,
  Omar Khattab, Peter Henderson, Qian Huang, Ryan~Andrew Chi, Sang~Michael Xie,
  Shibani Santurkar, Surya Ganguli, Tatsunori Hashimoto, Thomas Icard, Tianyi
  Zhang, Vishrav Chaudhary, William Wang, Xuechen Li, Yifan Mai, Yuhui Zhang,
  and Yuta Koreeda. 2023.
\newblock \href {https://openreview.net/forum?id=iO4LZibEqW} {Holistic
  evaluation of language models}.
\newblock \emph{Transactions on Machine Learning Research}.
\newblock Featured Certification, Expert Certification.

\bibitem[{Luo et~al.(2022)Luo, Xue, Xing, and Sun}]{luo2022prcbert}
Xianchang Luo, Yinxing Xue, Zhenchang Xing, and Jiamou Sun. 2022.
\newblock {PRCBERT: Prompt Learning for Requirement Classification using
  BERT-based Pretrained Language Models}.
\newblock In \emph{Proceedings of the 37th IEEE/ACM International Conference on
  Automated Software Engineering}, pages 1--13.

\bibitem[{Maas et~al.(2011)Maas, Daly, Pham, Huang, Ng, and Potts}]{IMDB}
Andrew~L. Maas, Raymond~E. Daly, Peter~T. Pham, Dan Huang, Andrew~Y. Ng, and
  Christopher Potts. 2011.
\newblock \href {https://aclanthology.org/P11-1015/} {Learning word vectors for
  sentiment analysis}.
\newblock In \emph{The 49th Annual Meeting of the Association for Computational
  Linguistics: Human Language Technologies, Proceedings of the Conference,
  19-24 June, 2011, Portland, Oregon, {USA}}, pages 142--150. The Association
  for Computer Linguistics.

\bibitem[{Malo et~al.(2014)Malo, Sinha, Korhonen, Wallenius, and
  Takala}]{finance}
Pekka Malo, Ankur Sinha, Pekka~J. Korhonen, Jyrki Wallenius, and Pyry Takala.
  2014.
\newblock \href {https://doi.org/10.1002/ASI.23062} {Good debt or bad debt:
  Detecting semantic orientations in economic texts}.
\newblock \emph{J. Assoc. Inf. Sci. Technol.}, 65(4):782--796.

\bibitem[{Mendes et~al.(2020)Mendes, Gon\c{c}alves, Cunha, Rocha, Couto-Rosa,
  and Martins}]{mendes2020keep}
Luiz~Felipe Mendes, Marcos Gon\c{c}alves, Washington Cunha, Leonardo Rocha,
  Thierson Couto-Rosa, and Wellington Martins. 2020.
\newblock \href {https://doi.org/10.1145/3340531.3412180} {"keep it simple,
  lazy" -- metalazy: A new metastrategy for lazy text classification}.
\newblock In \emph{Proceedings of the 29th ACM International Conference on
  Information \& Knowledge Management}, CIKM '20, page 1125–1134, New York,
  NY, USA. Association for Computing Machinery.

\bibitem[{Pang and Lee(2005)}]{PangMovie}
Bo~Pang and Lillian Lee. 2005.
\newblock \href {https://doi.org/10.3115/1219840.1219855} {Seeing stars:
  Exploiting class relationships for sentiment categorization with respect to
  rating scales}.
\newblock In \emph{Proceedings of the 43rd Annual Meeting of the Association
  for Computational Linguistics ({ACL}{'}05)}, pages 115--124, Ann Arbor,
  Michigan. Association for Computational Linguistics.

\bibitem[{Pasin et~al.(2024)Pasin, Cunha, Gon{\c{c}}alves, and
  Ferro}]{pasin2024quantum}
Andrea Pasin, Washington Cunha, Marcos~Andr{\'e} Gon{\c{c}}alves, and Nicola
  Ferro. 2024.
\newblock A quantum annealing instance selection approach for efficient and
  effective transformer fine-tuning.
\newblock In \emph{Proceedings of the 2024 ACM SIGIR International Conference
  on Theory of Information Retrieval}, pages 205--214.

\bibitem[{Penedo et~al.(2023)Penedo, Malartic, Hesslow, Cojocaru, Cappelli,
  Alobeidli, Pannier, Almazrouei, and Launay}]{falcon}
Guilherme Penedo, Quentin Malartic, Daniel Hesslow, Ruxandra Cojocaru,
  Alessandro Cappelli, Hamza Alobeidli, Baptiste Pannier, Ebtesam Almazrouei,
  and Julien Launay. 2023.
\newblock \href {https://doi.org/10.48550/ARXIV.2306.01116} {The refinedweb
  dataset for falcon {LLM:} outperforming curated corpora with web data, and
  web data only}.
\newblock \emph{CoRR}, abs/2306.01116.

\bibitem[{Rosenthal et~al.(2019)Rosenthal, Farra, and Nakov}]{semEval2017}
Sara Rosenthal, Noura Farra, and Preslav Nakov. 2019.
\newblock \href {http://arxiv.org/abs/1912.00741} {Semeval-2017 task 4:
  Sentiment analysis in twitter}.
\newblock \emph{CoRR}, abs/1912.00741.

\bibitem[{Socher et~al.(2013)Socher, Perelygin, Wu, Chuang, Manning, Ng, and
  Potts}]{sst2}
Richard Socher, Alex Perelygin, Jean Wu, Jason Chuang, Christopher~D. Manning,
  Andrew~Y. Ng, and Christopher Potts. 2013.
\newblock \href {https://aclanthology.org/D13-1170/} {Recursive deep models for
  semantic compositionality over a sentiment treebank}.
\newblock In \emph{Proceedings of the 2013 Conference on Empirical Methods in
  Natural Language Processing, {EMNLP} 2013, 18-21 October 2013, Grand Hyatt
  Seattle, Seattle, Washington, USA, {A} meeting of SIGDAT, a Special Interest
  Group of the {ACL}}, pages 1631--1642. {ACL}.

\bibitem[{Sokolova and Lapalme(2009)}]{Sokolova}
Marina Sokolova and Guy Lapalme. 2009.
\newblock \href {https://doi.org/10.1016/j.ipm.2009.03.002} {A systematic
  analysis of performance measures for classification tasks}.
\newblock \emph{Information Processing \& Management ({IP\&M})},
  45(4):427--437.

\bibitem[{Sorensen et~al.(2022)Sorensen, Robinson, Rytting, Shaw, Rogers,
  Delorey, Khalil, Fulda, and Wingate}]{avalia_varios_prompts}
Taylor Sorensen, Joshua Robinson, Christopher~Michael Rytting, Alexander~Glenn
  Shaw, Kyle~Jeffrey Rogers, Alexia~Pauline Delorey, Mahmoud Khalil, Nancy
  Fulda, and David Wingate. 2022.
\newblock \href {https://doi.org/10.18653/V1/2022.ACL-LONG.60} {An
  information-theoretic approach to prompt engineering without ground truth
  labels}.
\newblock In \emph{Proceedings of the 60th Annual Meeting of the Association
  for Computational Linguistics (Volume 1: Long Papers), {ACL} 2022, Dublin,
  Ireland, May 22-27, 2022}, pages 819--862. Association for Computational
  Linguistics.

\bibitem[{Spirling(2023)}]{spirling2023open}
Arthur Spirling. 2023.
\newblock Why open-source generative ai models are an ethical way forward for
  science.
\newblock \emph{Nature}, 616(7957):413--413.

\bibitem[{Strubell et~al.(2019)Strubell, Ganesh, and
  McCallum}]{strubell2019energy}
Emma Strubell, Ananya Ganesh, and Andrew McCallum. 2019.
\newblock Energy and policy considerations for deep learning in nlp.
\newblock \emph{arXiv preprint arXiv:1906.02243}.

\bibitem[{Touvron et~al.(2023)Touvron, Martin, Stone, Albert, Almahairi,
  Babaei, Bashlykov, Batra, Bhargava, Bhosale, Bikel, Blecher, Canton{-}Ferrer,
  Chen, Cucurull, Esiobu, Fernandes, Fu, Fu, Fuller, Gao, Goswami, Goyal,
  Hartshorn, Hosseini, Hou, Inan, Kardas, Kerkez, Khabsa, Kloumann, Korenev,
  Koura, Lachaux, Lavril, Lee, Liskovich, Lu, Mao, Martinet, Mihaylov, Mishra,
  Molybog, Nie, Poulton, Reizenstein, Rungta, Saladi, Schelten, Silva, Smith,
  Subramanian, Tan, Tang, Taylor, Williams, Kuan, Xu, Yan, Zarov, Zhang, Fan,
  Kambadur, Narang, Rodriguez, Stojnic, Edunov, and Scialom}]{llama2}
Hugo Touvron, Louis Martin, Kevin Stone, Peter Albert, Amjad Almahairi, Yasmine
  Babaei, Nikolay Bashlykov, Soumya Batra, Prajjwal Bhargava, Shruti Bhosale,
  Dan Bikel, Lukas Blecher, Cristian Canton{-}Ferrer, Moya Chen, Guillem
  Cucurull, David Esiobu, Jude Fernandes, Jeremy Fu, Wenyin Fu, Brian Fuller,
  Cynthia Gao, Vedanuj Goswami, Naman Goyal, Anthony Hartshorn, Saghar
  Hosseini, Rui Hou, Hakan Inan, Marcin Kardas, Viktor Kerkez, Madian Khabsa,
  Isabel Kloumann, Artem Korenev, Punit~Singh Koura, Marie{-}Anne Lachaux,
  Thibaut Lavril, Jenya Lee, Diana Liskovich, Yinghai Lu, Yuning Mao, Xavier
  Martinet, Todor Mihaylov, Pushkar Mishra, Igor Molybog, Yixin Nie, Andrew
  Poulton, Jeremy Reizenstein, Rashi Rungta, Kalyan Saladi, Alan Schelten, Ruan
  Silva, Eric~Michael Smith, Ranjan Subramanian, Xiaoqing~Ellen Tan, Binh Tang,
  Ross Taylor, Adina Williams, Jian~Xiang Kuan, Puxin Xu, Zheng Yan, Iliyan
  Zarov, Yuchen Zhang, Angela Fan, Melanie Kambadur, Sharan Narang,
  Aur{\'{e}}lien Rodriguez, Robert Stojnic, Sergey Edunov, and Thomas Scialom.
  2023.
\newblock \href {https://doi.org/10.48550/ARXIV.2307.09288} {Llama 2: Open
  foundation and fine-tuned chat models}.
\newblock \emph{CoRR}, abs/2307.09288.

\bibitem[{Viegas et~al.(2023)Viegas, Canuto, Cunha, Fran\c{c}a, Valiense,
  Rocha, and Gon\c{c}alves}]{yelp_viegas}
Felipe Viegas, Sergio Canuto, Washington Cunha, Celso Fran\c{c}a, Claudio
  Valiense, Leonardo Rocha, and Marcos~Andr\'{e} Gon\c{c}alves. 2023.
\newblock \href {https://doi.org/10.1145/3617023.3617039} {Clusent –
  combining semantic expansion and de-noising for dataset-oriented sentiment
  analysis of short texts}.
\newblock In \emph{Proceedings of the 29th Brazilian Symposium on Multimedia
  and the Web}, WebMedia '23, page 110–118, New York, NY, USA. Association
  for Computing Machinery.

\bibitem[{Ye et~al.(2023)Ye, Huang, Liang, and Chi}]{info14050262}
Feiyang Ye, Liang Huang, Senjie Liang, and KaiKai Chi. 2023.
\newblock \href {https://doi.org/10.3390/info14050262} {{Decomposed Two-Stage
  Prompt Learning for Few-Shot Named Entity Recognition}}.
\newblock \emph{Information}, 14(5).

\bibitem[{Zanotto et~al.(2021)Zanotto, Beck~da Silva~Etges, Dal~Bosco, Cortes,
  Ruschel, De~Souza, Andrade, Viegas, Canuto, Luiz et~al.}]{zanotto2021stroke}
Bruna~Stella Zanotto, Ana~Paula Beck~da Silva~Etges, Avner Dal~Bosco,
  Eduardo~Gabriel Cortes, Renata Ruschel, Ana~Claudia De~Souza, Claudio~MV
  Andrade, Felipe Viegas, Sergio Canuto, Washington Luiz, et~al. 2021.
\newblock Stroke outcome measurements from electronic medical records:
  cross-sectional study on the effectiveness of neural and nonneural
  classifiers.
\newblock \emph{JMIR Medical Informatics}, 9(11):e29120.

\end{thebibliography}
